\documentclass{article}

\usepackage{arxiv}

\usepackage[utf8]{inputenc} 
\usepackage[T1]{fontenc}    
\usepackage{hyperref}       
\usepackage{url}            
\usepackage{booktabs}       
\usepackage{amsfonts}       
\usepackage{nicefrac}       
\usepackage{microtype}      
\usepackage{lipsum}
\usepackage{graphicx}
\graphicspath{ {./images/} }

\usepackage{amsmath,amssymb,amsfonts}
\usepackage{algorithmic}
\usepackage[ruled,vlined,linesnumbered ]{algorithm2e}
\usepackage{graphicx}
\usepackage{graphicx,color}
\usepackage{comment}
\usepackage{soul}
\soulregister\cite7
\soulregister\ref7
\usepackage{xcolor,mdframed}

\usepackage{soul}
\usepackage{xifthen} 

\newif\ifhighlight

\newcommand{\maybehl}[1]{%
  \ifhighlight
    \hl{#1}%
  \else
    #1%
  \fi
}

  {\par\vspace{\abovedisplayskip}\noindent\begin{tabular}{l l l}}
  {\end{tabular}\par\vspace{\belowdisplayskip}}

\usepackage{multirow}
\usepackage{textcomp}
\def\BibTeX{{\rm B\kern-.05em{\sc i\kern-.025em b}\kern-.08em
    T\kern-.1667em\lower.7ex\hbox{E}\kern-.125emX}}
\AtBeginDocument{\definecolor{ojcolor}{cmyk}{0.93,0.59,0.15,0.02}}

\title{Comprehensive Assessment of LiDAR Evaluation Metrics: A Comparative Study Using Simulated and Real Data}

\author{
 Syed Mostaquim Ali \\
  National Research Council Canada \\
  London, ON, Canada\\
  Civil and Environmental Engineering\\
  Western University, London, ON, Canada \\
  \texttt{syedmostaquim.ali@nrc-cnrc.gc.ca;sali632@uwo.ca} \\
   \And
 Taufiq Rahman \\
  National Research Council Canada \\
  London, ON, Canada\\
  \texttt{taufiq.rahman@nrc-cnrc.gc.ca} \\
  \And
Ghazal Farhani \\
  National Research Council Canada \\
  London, ON, Canada\\
  \texttt{ghazal.farhani@nrc-cnrc.gc.ca} \\
  \And
  Mohamed H. Zaki \\
  Civil and Environmental Engineering\\
  Western University, London, ON, Canada \\
  \texttt{m.zaki@uwo.ca} \\
  \And
  Benoit Anctil \\
  Transport Canada \\
  Ottawa, Canada\\
  \texttt{benoit.antctil@tc.gc.ca}
  \And
  Dominique Charlebois\\
  Transport Canada \\
  Ottawa, Canada\\
  \texttt{dominique.charlebois@tc.gc.ca}
  \And
}

\begin{document}
\maketitle

\begin{abstract}
For developing safe Autonomous Driving Systems (ADS), rigorous testing is required before they are deemed safe for road deployments. Since comprehensive conventional physical testing is impractical due to cost and safety concerns, Virtual Testing Environments (VTE) can be adopted as an alternative. Comparing VTE-generated sensor outputs against their real-world analogues can be a strong indication that the VTE accurately represents reality. Correspondingly, this work explores a comprehensive experimental approach to finding evaluation metrics suitable for comparing real-world and simulated LiDAR scans. The metrics were tested in terms of sensitivity and accuracy with different noise, density, distortion, sensor orientation, and channel settings. From comparing the metrics, we found that Density Aware Chamfer Distance (DCD) works best across all cases. In the second step of the research, a Virtual Testing Environment was generated using real LiDAR scan data. The data was collected in a controlled environment with only static objects using an instrumented vehicle equipped with LiDAR, IMU and cameras. Simulated LiDAR scans were generated from the VTEs using the same pose as real LiDAR scans. The simulated and LiDAR scans were compared in terms of model perception and geometric similarity. Actual and simulated LiDAR scans have a similar semantic segmentation output with a mIoU of 21\% with corrected intensity and an average density aware chamfer distance (DCD) of 0.63. This indicates a slight difference in the geometric properties of simulated and real LiDAR scans and a significant difference between model outputs. During the comparison, density-aware chamfer distance was found to be the most correlated among the metrics with perception methods. 
\end{abstract}


\section{Introduction}
Autonomous driving systems (ADS) are significantly different from traditional vehicles, which are composed of mostly mechanical components. Even the few software components used in traditional vehicles (e.g., anti-lock braking system control, stability control, etc.) are developed from classical control theory to operate in a deterministic manner. In contrast, the software components in an ADS are much more extensive. Some algorithms, such as those that are machine learning based, are stochastic in nature.  Unlike traditional automotive software components, a causality based relationship between inputs and outputs cannot always be established. Furthermore, they are expected to operate in non-deterministic environments characterized by infinite numbers of permutations and combinations of different factors involving the operating environment, which renders obtaining absolute proof of safety impossible \cite{riedmaier2020survey}. In addition, traditional testing methods used for safety evaluation of automotive systems cannot be adapted for ADS \cite{every2017novel}. Unsurprisingly, it was estimated in \cite{kalra2016driving} that an ADS system must be driven 11 billion miles requiring 500 years of real-world operation just to prove with acceptable statistical significance that its capabilities are 20\% better than the safety benchmark of the average human driver of 1.09 fatalities per 100 million miles. Therefore, it is deemed impractical to rely on on-road and track testing alone to generate the test coverage needed for ADS. A feasible alternative is virtual testing \cite{zhang2021test} which can be used to generate effective test coverage at a much lower time and resource cost. 

Regulatory bodies such as the United Nations Economic Commission for Europe (UNECE) and the US Department of Transportation (USDOT) have already recognized virtual testing as an ADS validation methodology to complement track testing, real-world testing, in-service monitoring and reporting \cite{unece29,blanco2020fmvss}. However, many open questions must be answered first before virtual testing can be adopted as a regulatory tool. Some of these questions have been identified in literature \cite{blanco2020fmvss,ebert2019validation}: (a) how reliability is defined in relation to autonomous operation, (b) the minimum degree of fidelity of the simulation models to enable virtual testing, (c) how the simulation models are validated, and (d) logistical considerations such as what models are used, who supplies these models and how or if proprietary algorithms may be supplied. The scope of this paper will focus on many aspects of (a), (b) and (c) of the above mentioned questions in relation to the virtual testing environments (VTE). 

A VTE serves as a replacement for resource-intensive real-world testing. It includes models that describe the geometry, appearance and behavior of static (e.g., road lanes, traffic signs, buildings) and dynamic (e.g., vehicular traffic, pedestrians, weather) elements of the roadway environment, and the software that evaluates these models to provide a view of how the driving environment evolves in time and how the VTE under test performs in different test cases. Since the concept of operational design domain (ODD) \cite{saej3016} is used to determine the safety boundary of an ADS or feature thereof, it is expected that a VTE also encapsulates the ODD of the ADS under test. A VTE should closely mimic the defined ODD and any deviation should be statistically comparable. Therefore it is essential to define evaluation metrics to establish confidence for virtual testing. In this paper, available literature on evaluating high-fidelity 3D virtual testing environments for ADS is reviewed with a view to discuss the metrics and benchmarks proposed for evaluating VTEs. We aim to provide a comprehensive overview that can guide future research in establishing confidence for VTEs for ADS.

\section{Literature Review}
\subsection{LiDAR Sensing Technology}
LiDAR and cameras are two critical technologies for environmental perception in ADS technologies. Although cameras provide high-resolution images and can read traffic signs and lights, they are more susceptible to environmental factors like fog and rain. LiDAR sensor's ability to function in diverse lighting and weather conditions makes it a more suitable option than camera-based systems for autonomous vehicle developers \cite{ Rane2022LiDARBS}. Although LiDAR is more expensive than cameras, it provides more accurate spatial representation than cameras \cite{Wang2021ResearchOC}. It is a crucial sensing system for autonomous vehicle technology, particularly for L3 and higher-level capabilities \cite{Zhao2019RecentDO}. The sensor uses pulsed Laser waves to provide precise spatial data for localization and tracking \cite{Elhousni2020ASO}. With the advancement of LiDAR technologies, such as solid-state sensors, there is expected to be a reduction in LiDAR costs \cite{ Rane2022LiDARBS}. The ADS development and research community has been developing perception tools based on multi-object detection \cite{wang2023dsvt,liu2024lion, shi2022pillarnet} and semantic segmentation \cite{wu2024point,peng2024oacnns}. State-of-the-art deep earning techniques such as point transformers \cite{zhao2021point} and sparse convolution networks \cite{liu2015sparse} have significantly improved model perceptions for LiDAR point clouds.

\subsection{Software}
Simulation software has been an integral part of the ADS research and development cycle since the early days. There are as many as 72 distinct simulators used in ADS research \cite{koroglu_ads_review}. Out of the simulators CARLA, Airsim, SUMO are one of the most popular. Besides conventional software, racing simulators and commercial games like GTA V have been used for ADS development \cite{wymann2000torcs, richter2016playing}. 
%

The article \cite{wmg-software-5} reviewed available simulation tools to find their advantages, disadvantages, and testing objectives based on different criteria such as fidelity, licensing, interface compatibility, access to data, sensor compatibility, graphics and physics realism, and development effort. Previous literature was surveyed and 5 main simulation software analyzed: CARLA, LGSIM, Gemini, SUMO, and OpenPASS. Sumo is primarily a microscopic simulator, and OpenPASS is a text-based traffic simulator. Carla is an open-source high-fidelity driving simulator based on an Unreal Engine. Although CARLA has a difficult installation and build process, it is most flexible because of its versatile and well-maintained Python API. CARLA also has advanced support for simulated sensors. LGSIM is another open-source simulation software that runs on the Unity engine. LGSIM has an easy-to-use GUI but it lacks a well-maintained Python API. Both LGSIM and CARLA are high-fidelity simulators that require high processing power. esmini is more lightweight compared to CARLA or LGSIM. The disadvantage of esmini is the lack of capability to simulate sensors. CARLA is often chosen simulation software for autonomous driving simulation because of graphics and physics realism, the ability to be customized, and support for sensor simulation.

\subsection{Credibility Assessment Framework}
Using a virtual testing environment (VTE) as an alternative to real-world testing scenarios requires the VTE to be validated. A pipeline for building trust in VTEs should include tools, models, operators, and documentation. The article \cite{credibiliy} provides guidelines for assessing the credibility of modelling \& simulation for VTEs and proposes a credibility assessment framework (CAF). The framework evaluates capability, accuracy, correctness, usability, and fit for purpose. The CAF process includes documentation, code, calculation verification, and sensitivity analysis for the VTE platform. The article provides guidelines for establishing credibility by focusing on management, the team's experience and expertise, input and output data, and the validation process. It also emphasizes the importance of proper documentation of the credibility assessment process. 

\subsection{LiDAR simulations}
The first type of LiDAR simulation relied on physics-based point cloud simulation in a 3D virtual world \cite{carla, gazebo,yue2018lidar, gschwandtner2011blensor}. The 3D virtual world is mainly developed by computer-aided design (CAD) objects, which only sometimes represent the complexity and details of the real world. To minimize this semantic complexity, real-world datasets have been utilized \cite{xia2018gibson}. 
The more accurate LiDAR simulation method relies on reconstructing real-world scenarios into 3D meshes and then running physics-based simulations \cite{fang2020augmented,manivasagam2020lidarsim}. Point cloud-based generative models are also used to develop virtual environments. Neural Radiance Field (Nerf) is also used for generating simulated LiDAR point clouds \cite{zhang2024nerf}. There are also hybrid methods of simulating LiDAR using hybrid methods \cite{guillard2022learning, attal2021torf, heiden2020physics,li2023pcgen}.
\subsection{Research Gap}
The current state of the LiDAR simulation method validates simulation using qualitative methods and based on the visual comparison method. The qualitative method may be useful but it does not produce a reliable benchmark for comparing different simulation methods. Also, using 3D perception methods for comparison highly depends on model performance. The accuracy of the perception method when comparing real and simulated LiDAR scans depends highly on the model and training data. Also, as most 3D perception methods are black box models, having similar output does not guarantee that both scans are similar. Also, training these models requires a considerable training time for accurate inferences. For this reason, an alternative method for assessing the similarity of real and simulated LiDAR scans has been proposed in the paper. Using geometric similarity methods can eliminate the model bias of 3D perception methods while also producing a benchmark to compare different simulation methods.

\section{Proposed Qualification Protocol}

The literature provides many general concepts of how VTEs should be qualified for the purpose of ADS validation. Given the enormity of the scope involving a general VTE qualification protocol, even in conceptual form, the individual articles were found to focus on certain aspects of VTE qualification; e.g., scenario generation and definition, validating software tool-chain for safety-critical systems, traffic simulation, vehicle dynamics modeling, etc. Because sensing and perception is a core ADS competency, this paper will focus on qualification of sensors in a VTE. 

The synthetic sensor data produced by sensor models can be used in end-to-end simulation testing wherein every aspect of the VTE is simulated or in a X-in-loop testing methodology. Since the physical processes of how sensors produce perception data (e.g., point cloud generation for LiDAR, image formation for cameras, inertial measurements for an IMU, etc.) are well understood, the most prevalent gap exists in the process of modeling how different ODD affects sensor performance. This gap will be addressed in this paper by specifically focusing on: (a) generation of a ground-truth driving environment, (b) generation of simulated sensor data under different ODDs, (c) development of comparison metrics, (e) development and demonstration how these comparison metrics can be applied for the purpose of qualifying the sensor models in a VTE.

\subsection{Generation of Ground-Truth Driving Environment}
A survey vehicle equipped with high definition perception sensors (e.g., LiDAR, cameras, IMU, GNSS-RTK) will be used to create a 3D map of a real-world driving environment. The driving environment will be represented by 3D point clouds captured by one or multiple LiDAR sensors. The textures for the different objects in the driving environment will be captured and recreated from calibrated camera data. The GNSS-RTK system will be used to geo-reference the different elements of the ground-truth driving environment.

\subsection{Generation of Simulated Sensor Data Under Different ODDs}
The ground-truth driving environment will be imported into the CARLA simulator and an ego-vehicle will be programmed to drive in it. The ego-vehicle will be equipped with different sensors (e.g., LiDAR, camera, radar) and the built-in models will be used to generate synthetic data. Subsequently, additional weather simulation models (e.g., simulation of LiDAR in adverse weather \cite{kilic2021lidar}, weather image translation \cite{kwak2021adverse}) will be augmented with the synthetic data to simulate sensor data in different ODDs.

\subsection{Development of Comparison Metrics}
Comparison metrics that quantify the similarity of simulated sensor data with respect to the ground-truth data will be used to evaluate the simulator's ability to produce realistic synthetic data. For point-cloud data, cloud matching techniques that accounts for both 3D geometry and the environmental context will be used. Image similarity measures can be used for the image data.

\section{Developing Evaluation Metrics}
Maximizing the similarity between real LiDAR scans and their simulated counterparts is necessary to minimize the sim2real gap. Consider we have a perfect simulator $F_{sim}$, which we use to simulate a LiDAR scan $F_{sim}(x_i) = Y_{sim}$, where $x_i$ represents all relevant variables from the real world. Under perfect conditions, both simulated and real LiDAR scans should be identical. However, since it is nearly impossible to develop perfect scenarios, there will always be discrepancies between simulated and real outputs. A similarity metric is necessary to measure the similarity between a simulator and its real-world counterpart. This section outlines the criteria for selecting evaluation metrics and assessing their performance, sensitivity, and validity as a similarity metric for LiDAR scans.

\subsection{Conditions for Evaluation metrics}
We assume any output of sensor $X$ results from a function $F$. The function $F$ can result from a real sensor under real-world conditions $x_i$ or a simulated sensor in a virtual testing environment under simulated conditions $x_i$. The sensor output is determined by the conditions presented $x_i$ in both real and simulated scenarios. The conditions $x_i$ can be different factors such as:
\begin{itemize} 
\item \textbf{Environmental Factors:} location of objects, weather conditions, and temperature.
 \item \textbf{Odometry:} Orientation, translation, rotation, and vehicle speed.
\item \textbf{System Factors:} Number of sensor channels, field of view, LiDAR hardware characteristics, and computational limitations. \end{itemize}


The similarity metrics for comparing LiDAR scans should satisfy the following conditions:
\begin{enumerate}
    \item \textbf{Self-consistency}: For any LiDAR scan $a$ and evaluation metric $f$, the method should produce the most accurate result when comparing a scan with itself. For instance, $f(a, a)$ should be precisely zero for distance functions, indicating zero distance between the same LiDAR scan.
    \item \textbf{Symmetry}: For any two scans $a$ and $b$, the evaluation metric should satisfy the similarity, such that $f(a,b) = f(b, a)$. This property ensures logical consistency across multiple comparisons.
    \item \textbf{Sensitivity}: The metric should be sensitive to scan variations. Any change in the similarity between two scans should be reflected in the evaluation metric's output. This ensures that the metric accurately captures differences in sensor outputs.
    \item \textbf{Computation Efficiency}: The evaluation metric should be computationally efficient with CPU time and memory, allowing fast evaluation of large datasets.
    \item \textbf{Well-defined range:} The evaluation metric should have a well-defined range that provides an interpretable scale for comparison. This allows for consistent comparison across different simulations and units.
\end{enumerate}

\subsection{Generating Cases to Evaluate Similarity Metrics}
To assess the metrics' sensitivity to identifying changes in LiDAR scans, we generated a series of cases using both real and simulated LiDAR data. These cases evaluate the metric's responsiveness to changes in LiDAR scan characteristics.

\subsubsection{Channel settings}
\begin{figure*}[!ht]
    \centering
    \includegraphics[width=\linewidth]{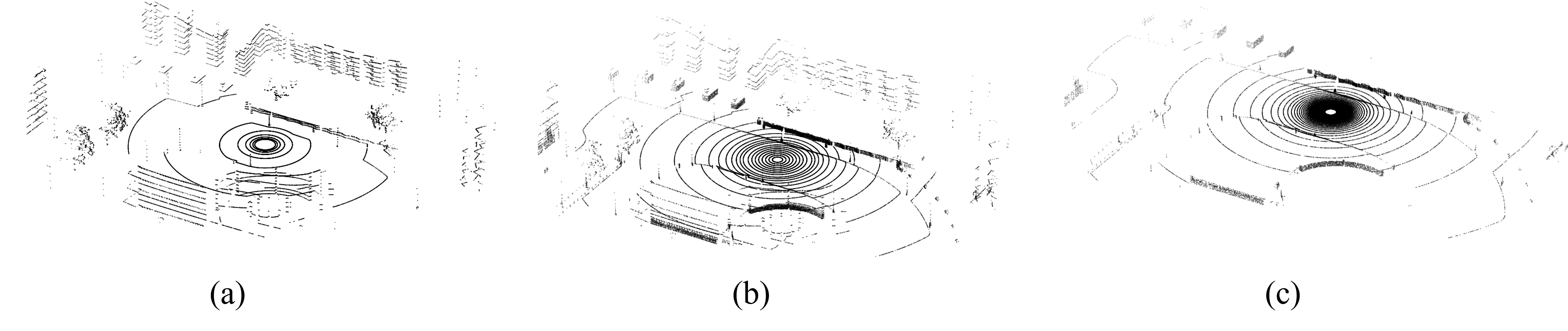}
    \caption{Simulated LiDAR scans of different channel settings from left VLP-16, VLP-32c, HDL-64E (top plots provide the spatial channel distribution for different LiDAR models)}
    \label{fig:sim_channels}
\end{figure*}

To evaluate the metrics response to different LiDAR settings, we simulate LiDAR scans in a Carla Environment using three widely used models: the 16-channel VLP-16, the 32-channel VLP-32c, and the 64-channel HDL-64E \maybehl{ which are commonly adopted in autonomous driving research \cite{geiger2012we}}. These different LiDAR sensors differ in \maybehl{the} number of points and vertical resolution. Figure \ref{fig:sim_channels} illustrates LiDAR scans generated with different simulated LiDARs. A Carla scene in an urban environment simulated these LiDAR scans with the same position and orientations. The objective of the experiment is to scan the same objects with exact positions and orientations with different LiDAR sensors. \maybehl{Furthermore, these scans will be used to determine the sensitivity of metrics to different LiDAR settings and the accuracy of metrics in identifying LiDAR scans of similar geometry, despite varying LiDAR settings.}

\subsubsection{Orientation}
 
\begin{figure}[!ht]
    \centering
    \includegraphics[width=\linewidth]{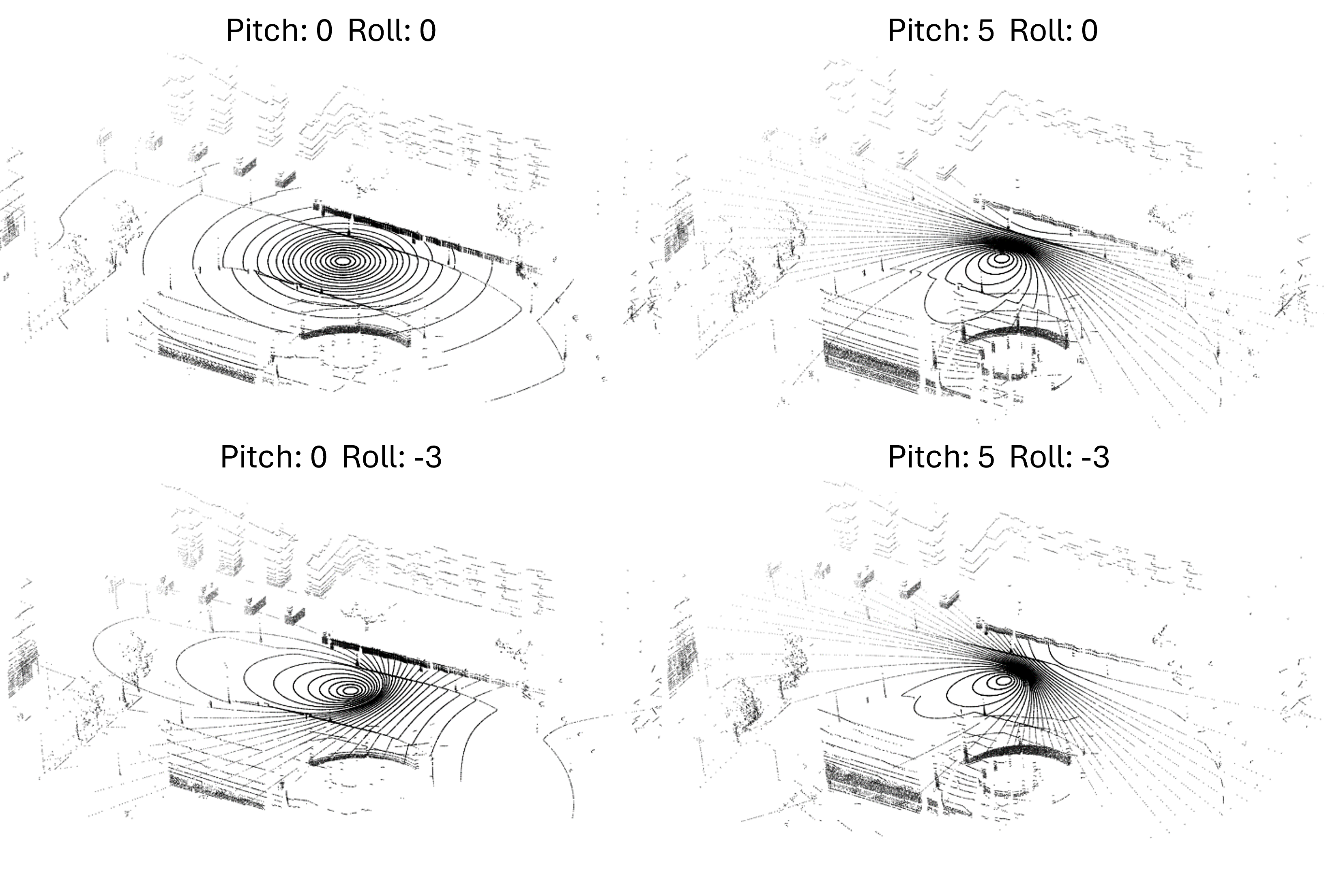}
    \caption{Simulated LiDAR scans with different orientations}
    \label{fig:sim_orientation}
\end{figure}

LiDAR scans with different orientations were simulated in a Carla Urban driving scene \maybehl{\cite{carla}}. This approach generates LiDAR scans with different orientations by simulating the rotation of the LiDAR in three axes. Simulated LiDAR scans were recorded in positions with different rotation values. This study was done in a simulated Environment, as replicating such conditions in real-world settings poses significant challenges.\maybehl{This experiment aims to assess the sensitivity of metrics to changes in orientation and determine whether metrics can accurately identify similar LiDAR scans of the exact location, despite different orientations.} Figure \ref{fig:sim_orientation} shows simulated LiDAR scans with different orientations.

\subsubsection{Distortion}
\begin{figure}[!ht]
    \centering
    \includegraphics[width=\linewidth]{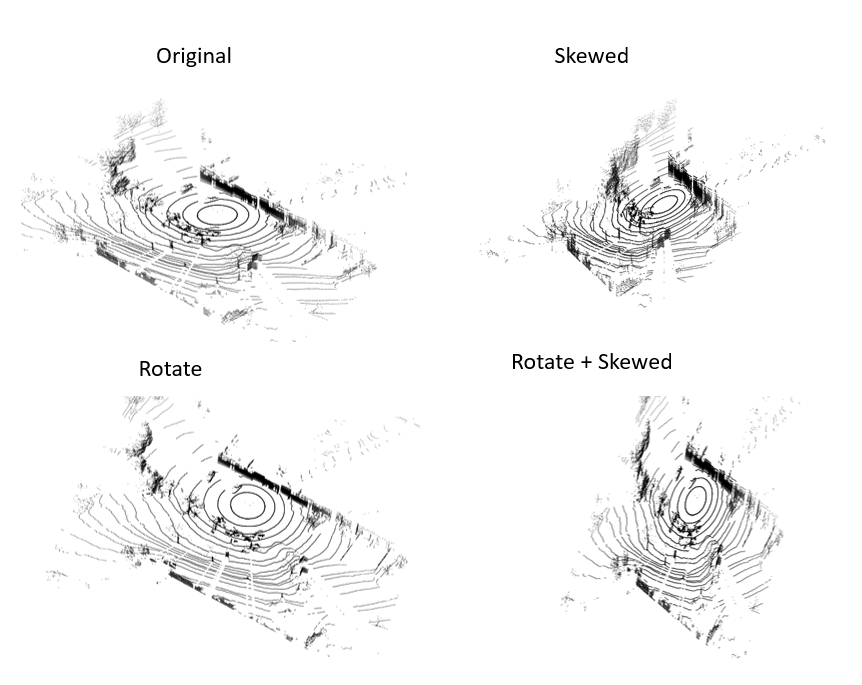}
    \caption{Simulated LiDAR scans with different distortions}
    \label{fig:distortion_sensitivity}
\end{figure}

Distortion in LiDAR scans can occur \maybehl{due to various factors, including the Doppler effect resulting from movement, vibrations, or inaccuracies in laser timing \cite{glennie2007rigorous}}. For our experiment with evaluation metrics, we introduced the distortion using spatial transformation. Real LiDAR scans are rotated, skewed, and translated to generate distorted and undistorted pairs for comparison. These pairs are compared through each evaluation metric to assess their performance in terms of sensitivity to distortion and accuracy of identifying original pairs. Figure \ref{fig:distortion_sensitivity} shows examples of cases of distortion.

\subsubsection{Noise}
\begin{figure}[!ht]
    \centering
    \includegraphics[width=0.9\linewidth]{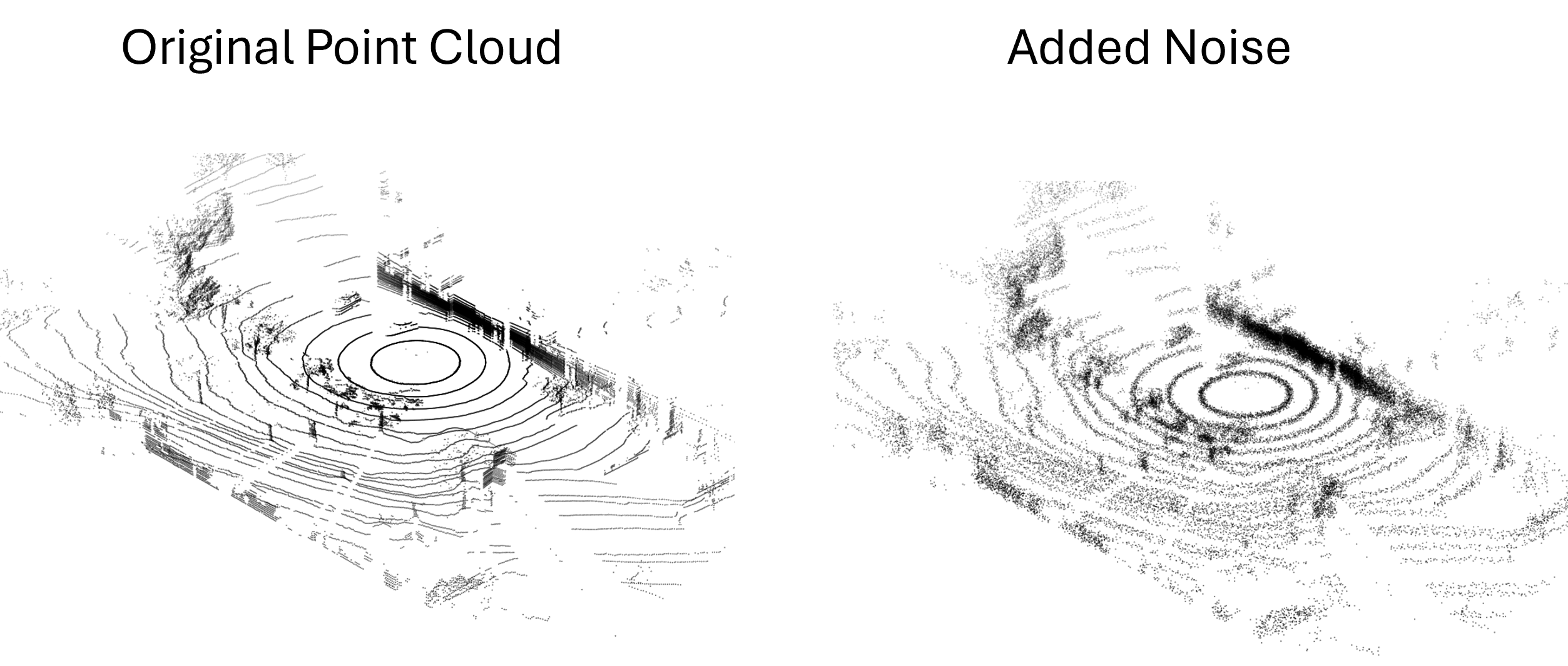}
    \caption{Added Noise to Real LiDAR Scans. In this example the magnitude of the noise is 0.2 normal standard deviation.}
    \label{fig:noise_sensitivity}
\end{figure}

Noise in LiDAR scans can occur due to \maybehl{ various factors, including weather conditions, sensor inaccuracies, or other environmental influences \cite{yan2020pointasnl}}. Random noise of varying intensities was introduced into the LiDAR scans. The noise intensity added was measured using standard deviation to ensure a controlled and measurable approach. Noises of varying intensities were added to real LiDAR scans to create multiple original and noisy scan pairs. Figure \ref{fig:noise_sensitivity} shows LiDAR scans with added noise,  Each scan pair was used as a metrics input to assess performance for noisy data. Generating multiple cases with varying noise intensities allows us to compare metric performance for different noise levels in a controlled manner.

\subsubsection{Outliers}
\begin{figure}[!ht]
    \centering
    \includegraphics[width=0.8\linewidth]{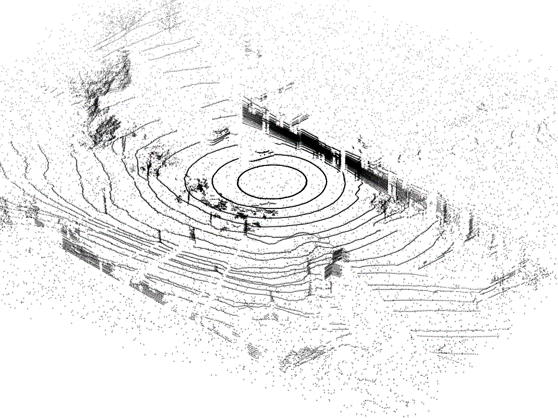}
    \caption{Added Random Scattered points into real LiDAR scans. The figure shows 10,000 random outlier points within the point cloud bounds.}
    \label{fig:outlier_points}
\end{figure}

\begin{figure}[!ht]
    \centering
    \includegraphics[width=0.8\linewidth]{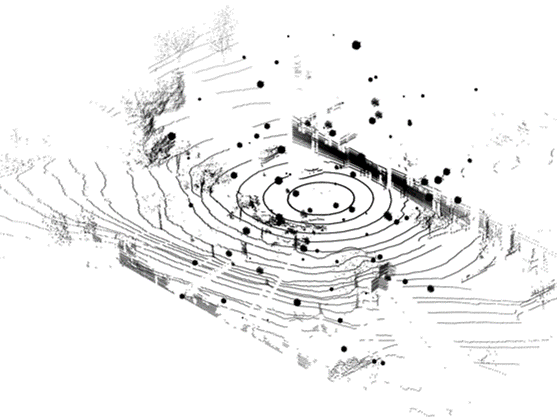}
    \caption{Added Clustered Outliers into real LiDAR scans. The figure shows 100 clustered points with random sizes with a limit of 1000 points. The clusters have a maximum radius of 1 unit. The clustered point clouds are to be added within a maximum of 20 units radius from the center of the LiDAR scan. }
    \label{fig:cluster_outliers}
\end{figure}

\maybehl{In real-world LiDAR systems, scans often contain outliers generated by physical, environmental, or sensor-related issues \cite{pirotti2018implementation}. Outliers are defined as points that do not conform to the geometric properties of the scanned scene. Two types of outliers usually occur in real-world scenarios. Random scattered outliers, which are isolated points, appear randomly across the scene, far from any surface in the scan. These often occur due to smoke, fog, dust or weather conditions such as snow or rain. Cluster scattered outliers are groups of erroneous points that do not correspond to any object in the scene. These are usually caused by reflections from glass, water or other reflective objects, and sensor errors.} Controlled outliers were introduced into real LiDAR scan point clouds to create test pairs for evaluation metrics. We added two outliers for our experiment: randomly scattered and clustered points. The random scattered points were distributed within the bounds of the LiDAR scan to simulate isolated outliers in the scan. The number of points measured the intensity of \maybehl{the} outliers added. Figure \ref{fig:outlier_points} shows a LiDAR scan with randomly scattered outliers. The clustered outliers consisted of points within a fixed radius and position. The clusters were added randomly to real LiDAR scans of varying positions, radii, and number of points within a threshold. Figure \ref{fig:cluster_outliers} shows a LiDAR scan with clustered outliers. Here, the number of clusters determined the intensity of the outliers. The approach of generating LiDAR scans with varying outliers allows for a detailed analysis of how different metrics are affected by outliers. 

\subsubsection{Density}

Point density is one of the critical properties of LiDAR scans \maybehl{affected by sensor type, distance to objects, and environmental factors \cite{ hackel2016fast}. } To evaluate the sensitivity of metrics to variations in point density, we generated LiDAR scans with lower density by downsampling real-world LiDAR data. From real LiDAR scans, a fraction of points \maybehl{is} randomly sampled to generate lower-density point clouds.
\maybehl{Random sampling is used instead of voxel-based down-sampling \cite{brock2016generative} because random sampling introduces an imbalance in point cloud density, whereas voxel-based down-sampling maintains a uniform distribution. Generating point clouds with varying densities enables us to test the metric's sensitivity to varying densities and its ability to identify original pairs.}


\subsection{Geometric Similarity Measures}
Simulating LiDAR scans should closely resemble their real-world counterparts in terms of their geometric properties. The similarity metrics proposed focus on finding geometric similarity to generate an evaluation score that quantifies the similarity between LiDAR scans. In this naive approach, we only measure the similarity of LiDAR scans in terms of geometry while ignoring other features, such as intensity. This method two LiDAR scans $P_1$ and $P_2$ with $n$ and $m$ number of points are compared where, 

\begin{itemize}
    \item  $P_1 = {p_1^{(1)}, p_1^{(2)}, \ldots p_1^{(n)}};$
    \item  $P_2 = {p_2^{(1)}, p_2^{(2)}, \ldots p_2^{(m)}};$
\end{itemize}
here, 
$p$ refers to each point in the pointcloud. The exponent refers to each unique point. 

\subsubsection{Chamfer Distance}
The chamfer distance \cite{butt1998optimum} is a commonly used method for point clouds in various applications. This method measures the minimum average Euclidean distance between two sets of points. The chamfer distance between two point clouds can be defined in equation \ref{eqn:chamfer_distance}. This method is symmetric because of its two components. It measures the average squared distance from each point $P_1$ to its nearest point in $P_2$. Then, in the second term for $P_2$ and $P_1$, vice versa. A smaller chamfer distance indicates greater similarity between point clouds, making it an excellent geometric similarity measure between two LiDAR data sets. The range of Chamfer distance is $0\to\infty$. Where 0 indicates the maximum similarity.

\begin{equation}
    \label{eqn:chamfer_distance}
    d_{chamfer}(P_1, P_2) = \frac{1}{|P_1|} \sum_{p_1\epsilon p_2}^{}||p_1-p_2||^2 +\frac{1}{|P_2|} \sum_{p_2\epsilon p_1}^{}||p_2-p_1||^2
\end{equation}

In the equation \ref{eqn:chamfer_distance}, $d_{chamfer}$ refers to the chamfer distance, $P$ refers to pointcloud. \maybehl{$p_1$ and $p_2$ refers to each point in the $P_1$ and $P_2$ pointclouds, respectively.} 
\subsubsection{Earth Mover's Distance}
The Earth Mover's Distance (EMD) is one of the most widely used methods for quantifying similarity between point clouds \cite{pele2009fast}. EMD calculates the minimum work required to transform one set of points into another. The earth mover distance is formulated as an optimization problem that finds a one-to-one bijection mapping, denoted as $\phi : P_1\hookrightarrow P_2$. The algorithm is presented in equation \ref{eqn:emd}, where the goal is to minimize the total cost of transforming the point cloud $P_1$ into $P_2$. Although this method provides a more detailed measure of dissimilarity than simple metrics like Chamfer Distance, this method requires both point clouds to be the same size. This property of EMD makes it impractical to compare LiDAR scans, as the number of point clouds generated in a LiDAR scan can not be guaranteed. 

\begin{equation}
    \label{eqn:emd}
     d_{\text{EMD}}(P_1, P_2) = \min_{\phi : P_1 \to P_2} \sum_{p_1\epsilon P_1} ||p_1 - \phi(p_1)||
\end{equation}
\ref{eqn:emd}, $d_{\text{EMD}}$ refers to the \maybehl{earth mover’s distance}, $P$ refers to pointcloud. \maybehl{$p_1$ and $p_2$ refers to each point in the $P_1$ and $P_2$ pointclouds, respectively.} . \maybehl{$\phi$} refers to the bijection mapping.

\subsubsection{Density Aware Chamfer Distance}
Wu et al. \cite{DCD} introduced Density Aware Chamfer Distance (DCD) as an advanced method for comparing point clouds. The DCD is calculated as shown in equation \ref{eqn:dcd}. The formulation of this method incorporates the term $e^z$, where $z<0$, ensuring the resulting distance is bounded between 0 and 1. The metric also introduces a scaling parameter for the distance $\alpha$, which can be used to adjust the variations in distances, allowing for a more precise comparison of point clouds. This method is more sensitive to disparities in point cloud than Chamfer distance and computationally efficient than Earth mover's distance. Also, the metric does not require a point cloud of similar sizes to compare, which makes it perfect as a geometric evaluation metric for LiDAR scans. 
\begin{equation}
\begin{aligned}
    \label{eqn:dcd}
    d_{\text{DCD}}(P_1, P_2) = \frac{1}{2} \left( \frac{1}{||P_1||} \sum_{p_1 \in P_1} \left(1 - \frac{1}{n_{\hat{p_1}}} e^{-\alpha ||p_1 - \hat{p_2}||_2} \right) + \right. \\
    \left. \frac{1}{||P_2||} \sum_{p_2 \in P_2} \left(1 - \frac{1}{n_{\hat{p_2}}} e^{-\alpha ||p_2 - \hat{p_1}||_2} \right) \right)
\end{aligned}
\end{equation}
\ref{eqn:dcd}, $d_{\text{DCD}}$ refers to the \maybehl{density aware chamfer distance}, $P$ refers to pointcloud. \maybehl{$p_1$ and $p_2$ refers to each point in the $P_1$ and $P_2$ pointclouds, respectively} , $\alpha$ refers to the sensitivty, and $n$ refers to number of points in the pointcloud. 
\subsubsection{Histogram Method}
This method converts the distance between points into a histogram \cite{Histogram-compare}. The histogram is generated by finding the pairwise distance between points from the object's surface. This method takes random point samples from the point cloud to control the run time. The histogram method is very suitable for unstructured point clouds.
Wallace et al. \cite{virtual-secnario-wmg, wallace2022validating} propose a custom histogram method for point clouds. In their method, distances are normalized by using the largest object distance. They claimed this method of normalization helps to preserve distance. The Minkowski distance was used to compare the absolute differences between the two histograms. Since this method calculates pairwise distances, the memory requirement is $O(n^2)$. This limitation means that to compare two point clouds with N points, it needs to compute a histogram from a matrix of size $2N^2$ elements. For this reason, the author suggested random sampling point clouds to N elements. We also tested the histogram method with Voxel downsampling to test the performance of metrics.

\subsubsection{Iterative Closest Point}
Iterative closest point is a point cloud registration method that registers two point clouds by finding the nearest point in the other cloud and calculating the distance between them. This method uses different objective functions to measure overlaps, such as point-to-point distance, point-to-plane distance, and surface normal alignment. Chamfer distance is one of the most popular optimization metrics used for the iterative closest point method. For our tests of geometric comparison, Iterative closest points with chamfer distance as a loss function were used as a similarity metric. The metric registers LiDAR scans together and generates a final inlier root mean square error between the registered point clouds. The final score is used as a metric for evaluation. One of the disadvantages of this method is that the Iterative Closest Point is not symmetric, and the registration process is computationally expensive.  

\section{Comparison of Geometric Similarity Metrics}
This section aims to discuss the capability of the similarity metrics. The similarity metrics are tested in different scenarios, and the results are discussed.

\subsection{Sensitivity of Metrics}




\subsubsection{Distortion: Rotation}
The metrics were tested with multiple LiDAR scans with different rotations to find the sensitivity of the metrics. The results show that Chamfer distance and Birds Eye View increase consistently with rotation. Density-aware Chamfer Distance and Voxel IoU are seen to be very sensitive to slight changes in rotation. The histogram method and the Iterative Closest point show the slightest sensitivity to rotation.


\subsubsection{Distortion: Translation}
The Chamfer distance and Bird's Eye View increase linearly with translation. The density-aware Chamfer distance and Voxel comparison are very sensitive to translation. The histogram approach is not sensitive to translation. The key takeaway from this comparison is that if the LiDAR scan has any translation distortion, it is required to find scans with Similar geometries. The histogram method is the metric for this purpose. 

\subsubsection{Distortion: Scale}
We see the metrics Chamfer Distance and Bird's eye view increase linearly when scaling up, but these metrics show the least distance for scaling down. Density aware chamfer distance, ICP (Iterative Closest Point) compare, voxel IoU, and histogram show maximum accuracy for scaling down and up. The Chamfer distance falsely identifies scaled-down LiDAR scans as similar LiDAR scans with zero distance, which might make chamfer distance impractical for scaled-down LiDAR scans.  


\subsubsection{Noise}

\begin{figure}[!ht]
    \centering
    \includegraphics[width=1\linewidth]{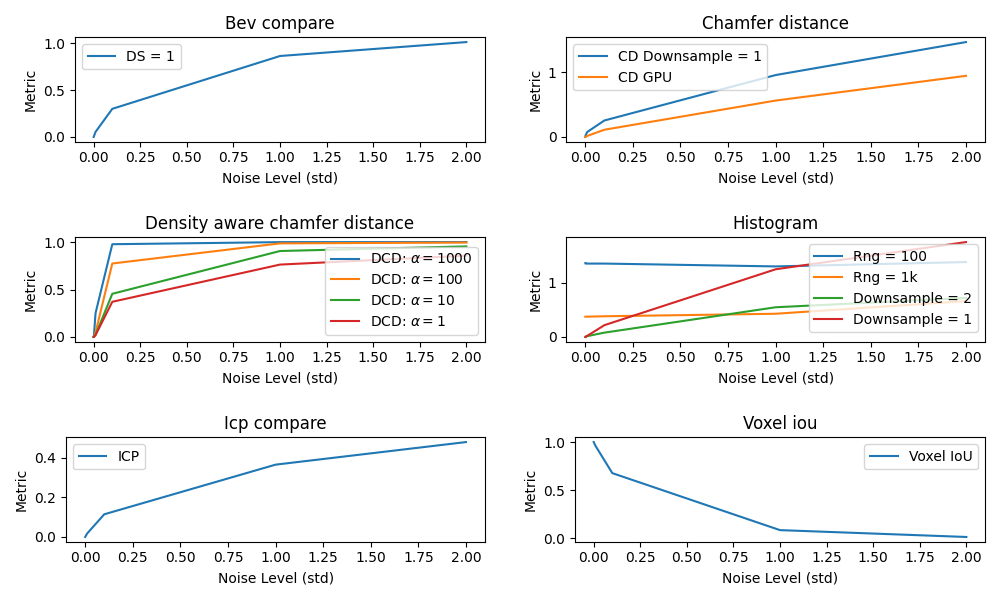}
    \caption{Metric Sensitivity to Added Noise}
    \label{fig:random_noise}
\end{figure}

The evaluation metrics were tested across different noise levels, up to 2 standard deviations. Figure \ref{fig:random_noise} shows the sensitivity of the metrics. The density-aware chamfer distance (DCD) demonstrated high sensitivity to noise levels. Decreasing the sensitivity $\alpha$ increases the metric DCD's sensitivity range. The Chamfer distance demonstrated a linear increase in response to noise and downsampling point cloud before comparing increases in sensitivity. Birds Eye View comparison and ICP comparison also showed a linear increase with noise. The Voxel IoU metric followed a similar trend. However, the histogram approach using random sampling showed minimal sampling to noise, although voxel downsampling with the histogram approach showed linear sensitivity toward noise levels. Overall, the results indicate that all the metrics, except for the histogram with random sampling, exhibit moderate sensitivity to noise

\subsubsection{Outliers}
\begin{figure}[!ht]
    \centering
    \includegraphics[width=1\linewidth]{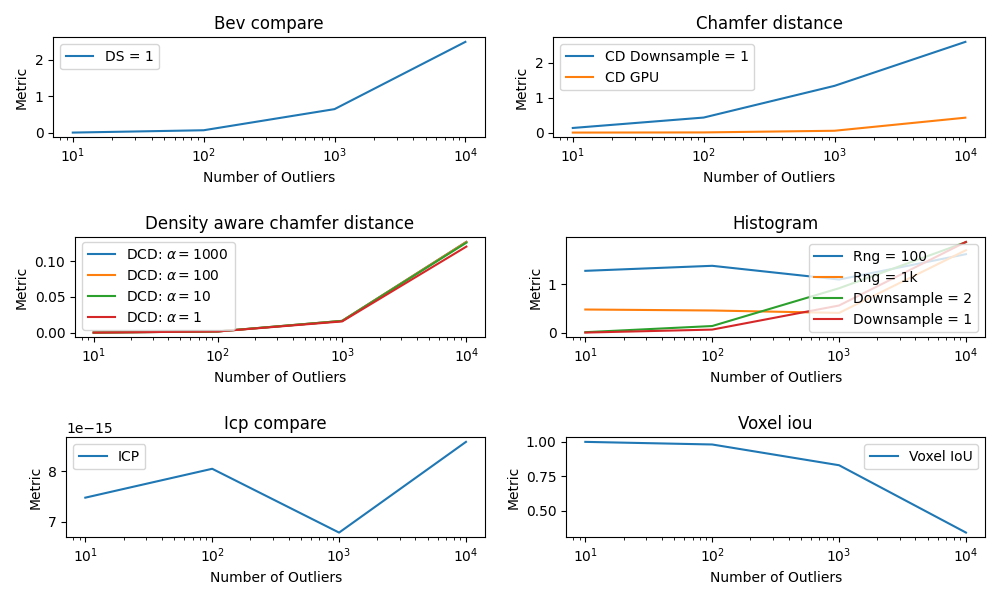}
    \caption{Metric Sensitivity to Random Outliers}
    \label{fig:random_outlier_result_graph}
\end{figure}

Testing outliers sensitivity we have two different cases. One where we compare the outlier sensitivity to adding random points. And another is for clustered points. 
\textbf{Adding random points: } For testing outliers, we tested by adding points up to 10,000. Figure \ref{fig:random_outlier_result_graph} shows the metric output with different numbers of points. Density-aware chamfer distance shows the most linear increase with the number of added points. The trend for this metric holds for all $\alpha$ values. Bird eye view comparison and chamfer distance also show a linear trend. Downsampling the point cloud before calculating chamfer distance seems to increase the sensitivity of the metric. The iterative closest point method does not show any sensitivity towards added outliers. The histogram approach shows a linear increase with added points across all variations, where downsampling the point cloud shows a similar trend to Chamfer distance.

\begin{figure}[!ht]
    \centering
    \includegraphics[width=1\linewidth]{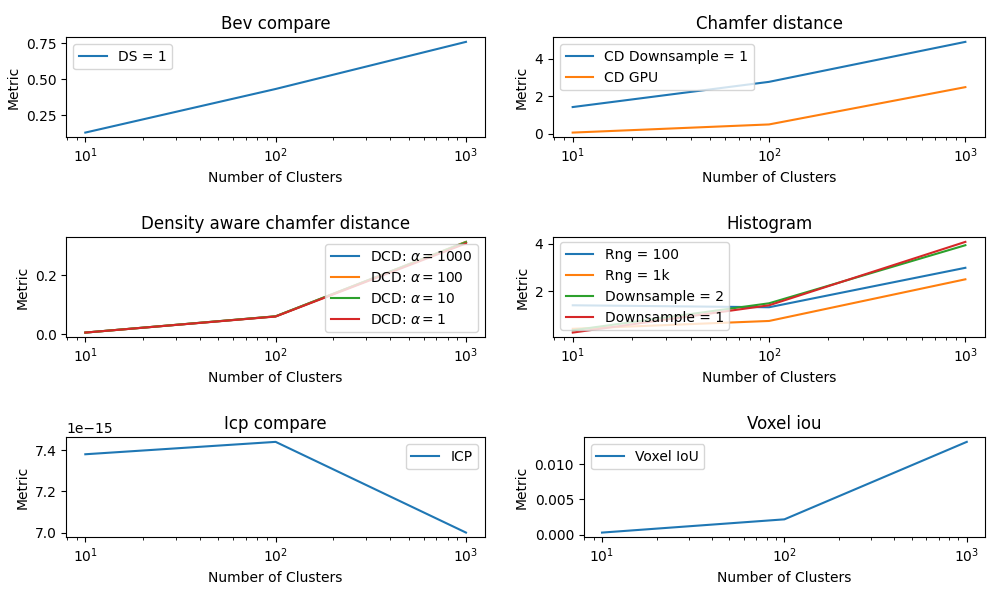}
    \caption{Metric Sensitivity to Clustered Outliers}
    \label{fig:clustered_outlier_result_graph}
\end{figure}

\textbf{Adding Clustered outliers: }
We tested the metrics comparison for point clouds with up to 1000 added clusters of a maximum size of 1000 points and 10 unit radius. In the figure \ref{fig:clustered_outlier_result_graph}, we see the output of metrics with different numbers of clusters. The comparison shows a linear increase for density-aware chamfer distance for all $\alpha$ values. Chamfer Distance shows a near linear increase. The ICP comparison and voxel method do not show sensitivity towards this approach. From the sensitivity analysis with outliers, it seems that all the methods are sensitive except for the ICP comparison method. Density-aware chamfer distance shows the most consistent increase with outliers.

\subsubsection{Density}
\begin{figure}[!ht]
    \centering
    \includegraphics[width=1\linewidth]{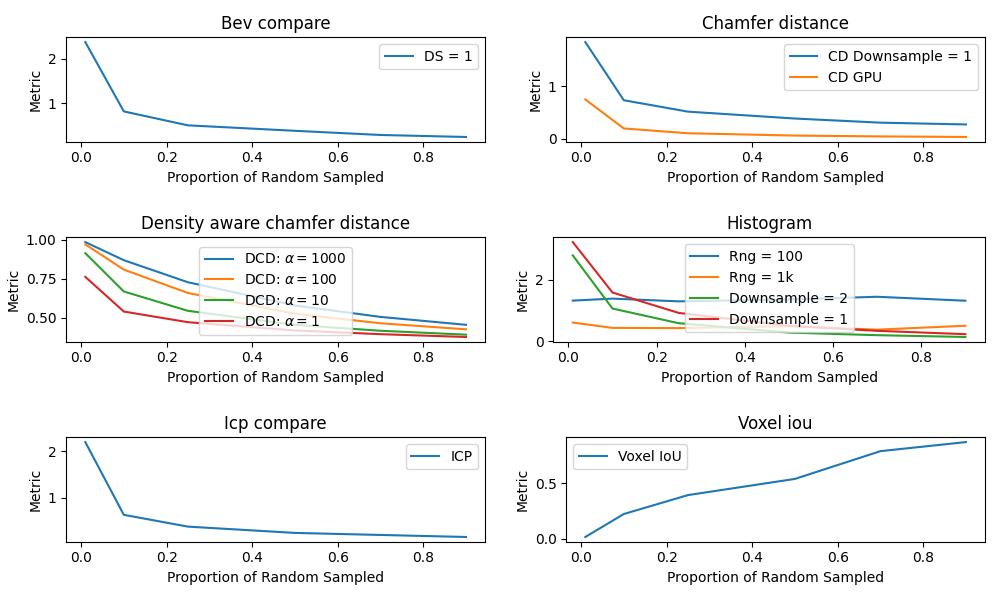}
    \caption{Metric Sensitivity to Density}
    \label{fig:density_result_graph}
\end{figure}

Lidar scans are randomly sampled to generate scans with different densities. Figure \ref{fig:density_result_graph} shows metric results with LiDAR scan sampled randomly with different proportions of the original scan. Chamfer distance shows moderate sensitivity, and gradually, the sensitivity decreases. Density-aware chamfer distance shows linear decreases for $\alpha=1000$. Bird's eye view, ICP, and voxel IoU show a similar trend as chamfer distance. The histogram method with random sampling is least sensitive to a density-based method, where downsampling shows a similar decrease as chamfer distance. From the analysis, density-aware chamfer distance shows the most linear increase with density proportion to the original scan.

\subsection{Accuracy of metrics}
\begin{figure*}
    \centering
    \includegraphics[width=1\linewidth]{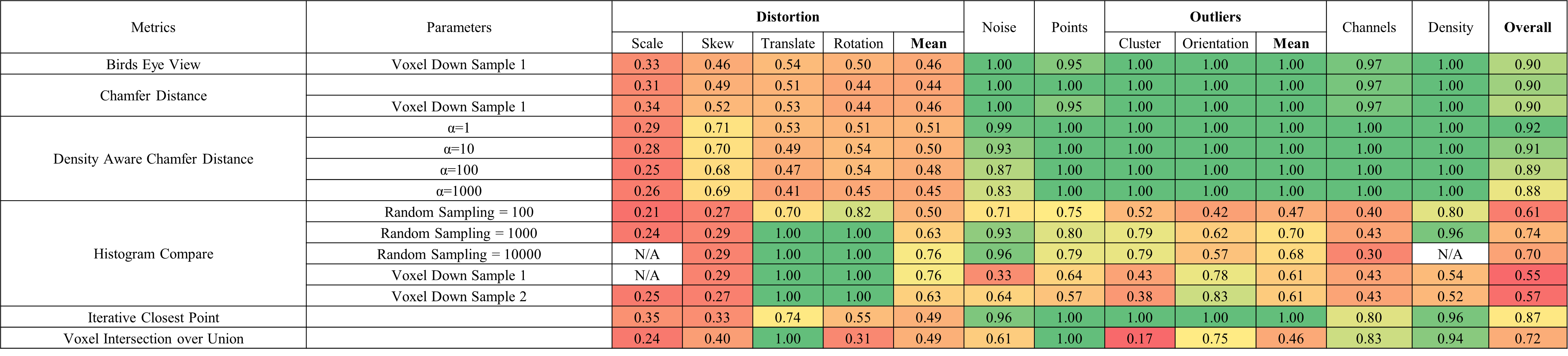}
    \caption{Accuracy of metrics when compared with different modifiers}
    \label{fig:accuracy-table}
\end{figure*}

LiDAR scans of the same positions were compared with each other with different modifiers (noise, outlier, distortion, orientation and channels). Each of the metrics was used to compare and generate an accuracy score. The accuracy score determines the metrics resilience to such modifications in LiDAR scans. The table \ref{fig:accuracy-table} shows the accuracy of different metrics. Out of all the metrics, the histogram method works best for translation and rotation scenarios without other distortions. From the table, we see Chamfer distance and density-aware Chamfer distance with $\alpha = 1$ worked best for all methods except for the distortion cases. None of the produced accurate results for Scale and Skew distortion scenarios. 

\subsection{Computation Time}
\begin{figure}[!ht]
    \centering
    \includegraphics[width=1\linewidth]{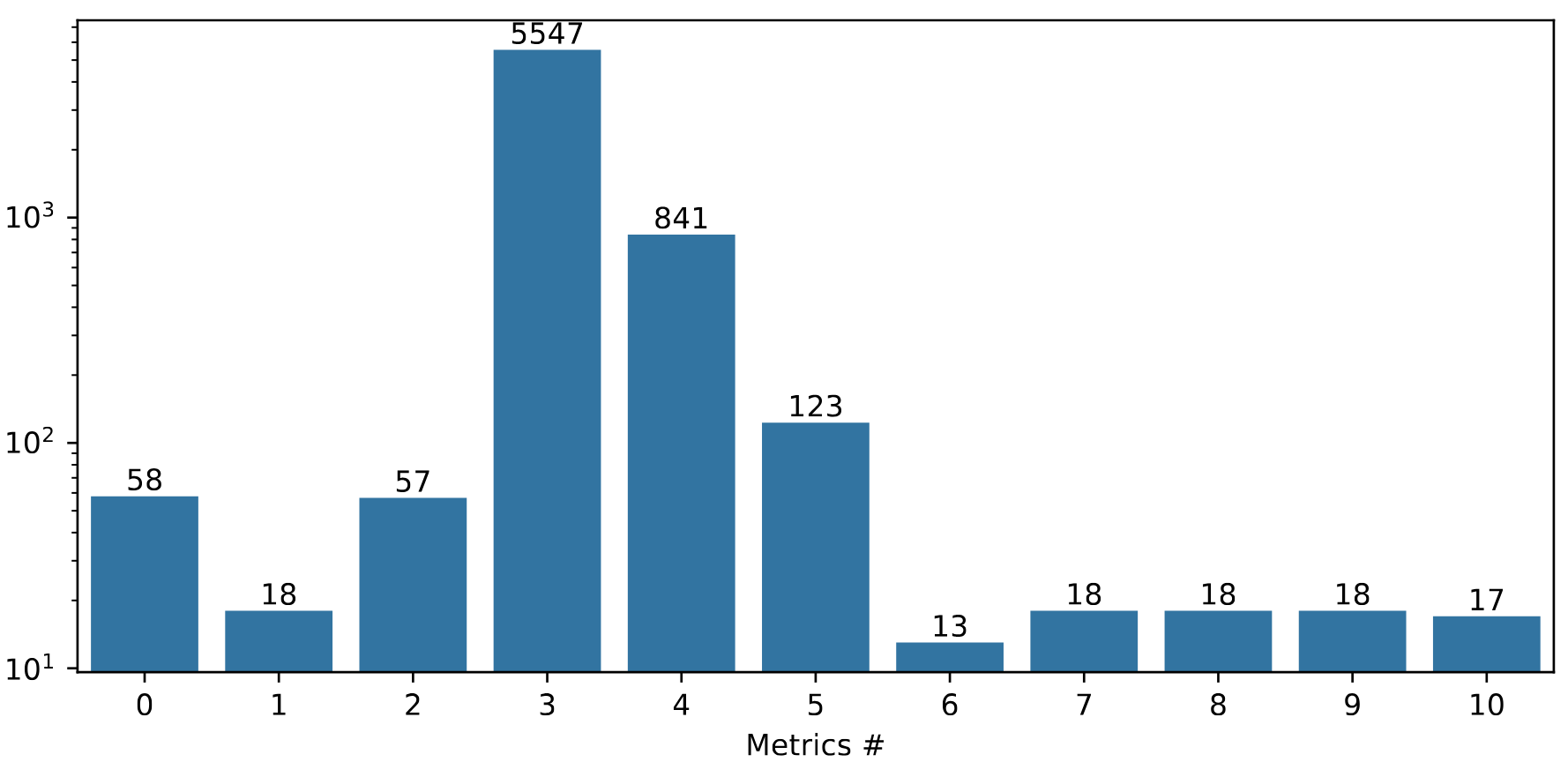}
    \caption{Computation Time. Birds Eye View (0), Chamfer Distance (1), Histogram  with Rng = 1000 points (2),with Rng =  10000 points (3),with Downsample = 1 (4), ICP (5), Voxel IoU (6), Density Aware Chamfer Distance; $\alpha = 1000$ (7), $\alpha = 100$ (8), $\alpha = 10$ (9), $\alpha = 1$ (10),     }
    \label{fig:computation-time-metric}
\end{figure}

The computation time for each metric is given in Figure \ref{fig:computation-time-metric}. The slowest evaluation metric is the histogram method with 10k points, followed by a bird's eye view without downsampling. The computation time for these metrics makes it impractical for large-scale use. Density-aware chamfer distance is the fastest approach among the computation metrics.

\subsection{Reflexivity of Metrics}
 The metrics $f$ compared random LiDAR scans $x$ with themselves to check if $f(a,a)$ should produce maximum similarity. All the metrics, except the histogram with random sampling, successfully output the maximum similarity. During our testing, we found that comparing the same point cloud with the metric histogram method with a random sampling of 100, 1000, and 10000 points produced values of $1.46 \pm 0.38$, $0.46 \pm .12$, and $.15 \pm 0.4$, respectively.

\subsection{Summary of Comparison}
From the Comparison, it is observed that each of the similar methods is well-suited for specific tasks. However, to find geometric similarity between real LiDAR scans, we aim to find a metric that shows linear sensitivity to deformation, has good accuracy, and has reasonable computation time. Chamfer Distance, Density Aware Chamfer Distance, and Bird's Eye View are suitable geometric similarity measures. Based on our analysis, we recommend a two-step approach for geometric similarity analysis. First, test with density-aware chamfer distance with $\alpha=1$. If the score is higher than 1, it means the scans are not very similar; testing again with chamfer distance will generate a better idea of how far real and simulated scans are. The combined approach optimizes accuracy, sensitivity, and computation efficiency.

\section{Virtual Test Environment Generation from Vehicle Data}
This section describes the digital twin development for the Carla simulator with collected real-world data. The data is collected with a vehicle mounted with LiDAR, IMU, and Camera sensors. These scans generate a 3D Carla scene using SLAM and 3D mesh generation techniques. The digital twin environment aims to create simulated replications of LiDAR scans collected in the real world. The LiDAR scans will then be evaluated to measure deviation from real-world scenarios. 


\subsection{Data Collection}
\begin{figure}[!ht]
    \centering
    \includegraphics[width=0.6\linewidth]{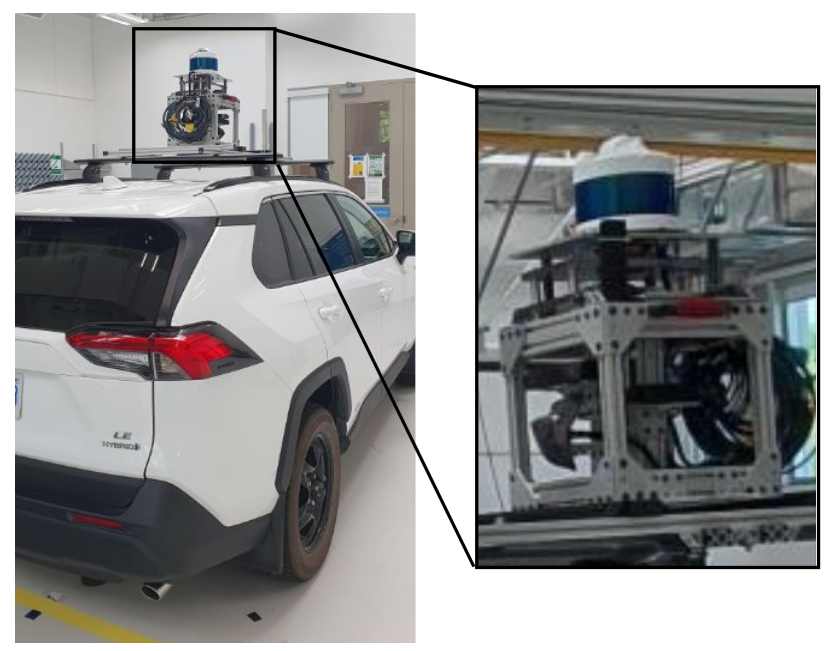}
    \caption{Data collection vehicle setup. }
    \label{fig:vehicle_setup}
\end{figure}

A 2021 Toyota RAV4 Hybrid is equipped with a Velodyne 32c LiDAR, 3 Zed 2i stereo depth camera, and an inertial measurement unit (IMU). The sensors were equipped with sufficient computing and power supply equipment. Figure \ref{fig:vehicle_setup} illustrates the vehicle setup. We employed the Robot Operating System (ROS) framework [41] for effective sensor management, deployment, and data handling. ROS provides a versatile environment for processing, recording, and analyzing sensor data through its modular node-based architecture. Communication within ROS occurs via nodes that exchange data as messages using a publisher-subscriber model. In this model, nodes publish messages to specific topics, and other nodes can subscribe to these topics to access the data. Additionally, ROS supports the recording of topic messages into ROS bag files, facilitating convenient offline analysis. During the survey, we recorded all sensor data into ROS bag files for subsequent detailed examination.
During the survey, the ROS framework recorded the messages for all sensors with accurate timestamps, allowing multi-sensor fusion. 

Western University Research Park was selected as the study area. The Figure shows the path taken by the vehicle. We strategically conducted the survey early in the morning, a time when the presence of pedestrians and cars was minimal. We aim to generate a digital twin to test LiDAR performance with static objects, as having dynamic objects in the scene may compromise the accuracy of digital twin generation. The vehicle covered a distance of 1 kilometre in driving, returning to its starting position. A closed-loop trajectory was required for the accuracy of the Simultaneous Localization and Mapping (SLAM) algorithm.



\subsubsection{Map Generation with SLAM}

LiDAR (Light Detection and Ranging) is a remote-sensing technology that measures a target's position by analyzing reflected light. Our experiment uses a 32-channel LiDAR VLP-32c, emitting laser pulses at 32 angles. This enables scanning multiple vertical planes, creating a detailed 3D point cloud. A single LiDAR scan provides an accurate 3D representation of points on targets from which lasers are reflected, providing a 3D snapshot of an area. The Simultaneous Localization and Mapping (SLAM) algorithm can be utilized to find the odometry of each LiDAR scan. This odometry information can be utilized to reconstruct the snapshots into a 3D point cloud map. The odometry of each scan describes the position $(t_x, t_y,t_z)$ and orientation $(r_x,r_y,r_z,r_w)$ of the sensor during data collection.

\begin{figure}
    \centering
    \includegraphics[width=1\linewidth]{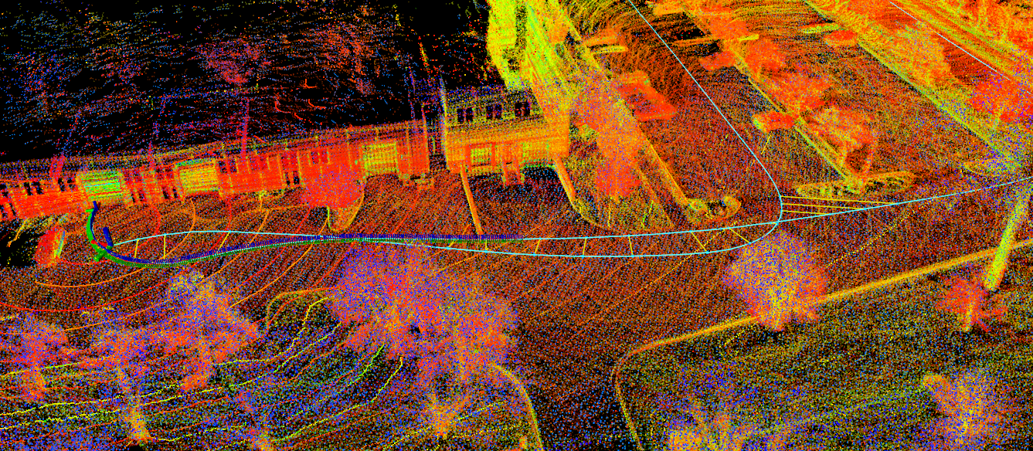}
    \caption{3D point-cloud reconstruction with LIO-SAM. The point colour shows the intensity of LiDAR scans. The green line indicates the trajectory of the vehicle.}
    \label{fig:liosam}
\end{figure}

This research utilizes LiDAR Inertial Odometry via Smooth Mapping (LIO-SAM) \cite{shan2020lio} for reconstructing 3D point clouds. This method integrates motion data from the Inertial Measurement Unit (IMU) and optimized odometry solution from LiDAR scans to generate accurate odometry of the LiDAR scans. The generated odometry is further corrected using loop closure methods. This fusion of multiple data enhances the precision of the generated 3d map. Figure \ref{fig:liosam} shows a snapshot of LiDAR reconstruction with SLAM. This process generates an accurate 3D map of the scanned area in meter coordinates. 

\subsubsection{Point Cloud Processing}
We use registered point clouds in conjunction with other environments and sensor information to generate the point cloud. Initially, the point cloud was generated in .ply format using the Python open3D library. The point cloud is a collection of points as each row contains position information $(x,y,z)$. We also add sensor position $(x,y,z)$ during recording into the point cloud. Using the normal estimation method, we use the sensor position to estimate the direction of normals for each point.




\subsubsection{Mesh Generation}
\begin{figure}[!ht]
\centering
\includegraphics[width= 0.7\linewidth]{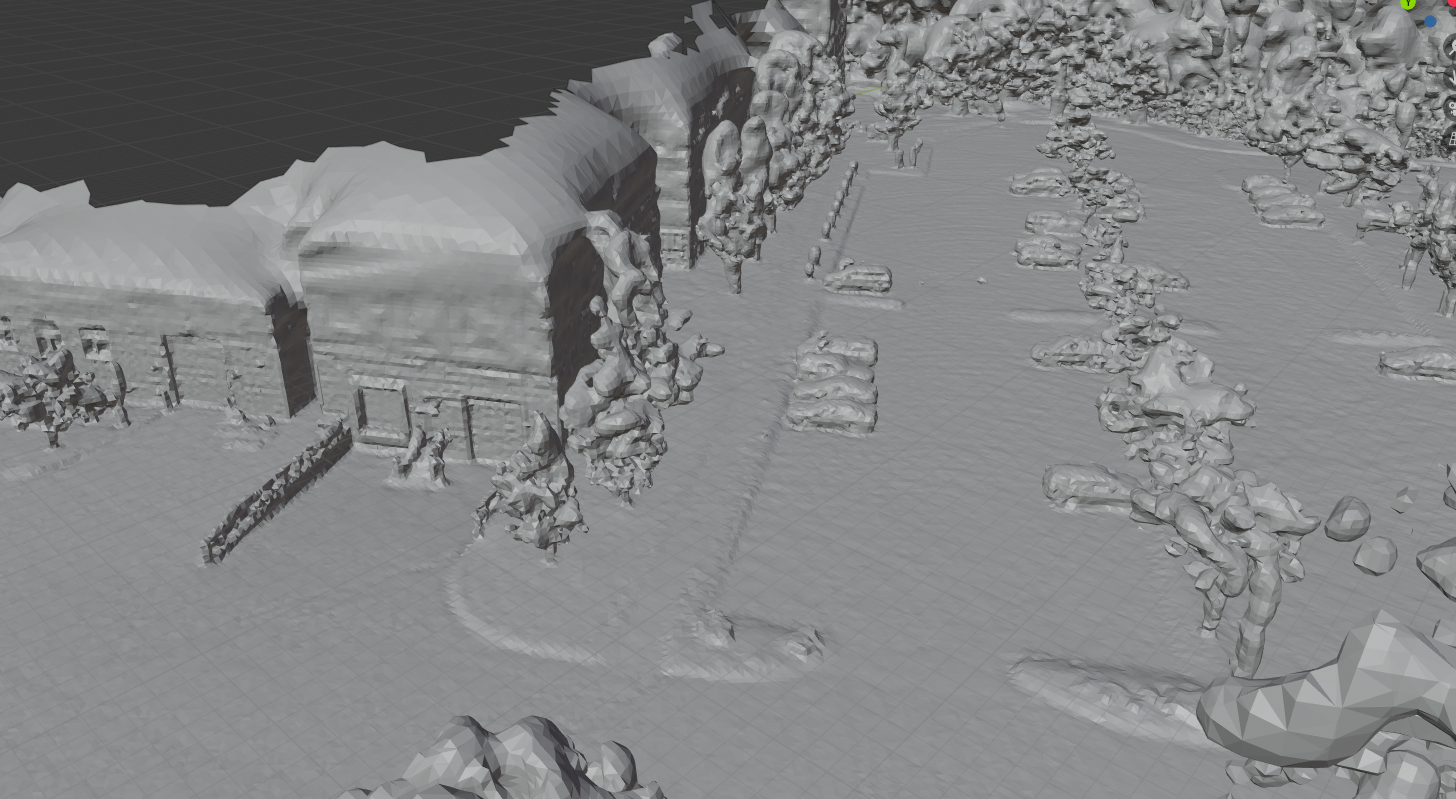}
\caption{Generated Mesh in Unreal Engine Environment.  }
\label{fig:mesh}
\end{figure}

Before generating the mesh, the point cloud was sub-sampled to reduce the number of points. Then, we generated the mesh from the subsampled point cloud using the Poisson reconstruction method. Initial faces on the mesh generated by this method are filtered based on the point density. Figure \ref{fig:mesh} shows the mesh generated. The mesh is then converted to FBX file format for importing to the unreal engine.

\subsubsection{Carla Scene Generation}
We used Carla's development build to develop the digital twin scenario. Generally, Carla requires a mesh and an open drive file to generate custom maps. However, as our goal was to generate a static scene for simulating sensors, we used an empty opendrive file. We created a blank project in Carla and imported the mesh. The original point cloud coordinate was in meters, so no additional scaling was required. The mesh's zero point origin was translated to the unreal engine's origin. The result is that the unreal coordinate matches 1:1 with the ROS coordinate. Since the objective of our research is LiDAR simulation, no additional texture or light adjustment was made for the scene. Figure \ref{fig:mesh} shows the scenario in Carla. 

\subsubsection{LiDAR Simulation}
Carla Ray cast LiDAR simulation, which was used to simulate LiDAR in a simulated environment. This method simulates rotating LiDAR, where each point is calculated using a single ray-casting vector for each channel. The rotation is simulated by capturing the horizontal angle at which the LiDAR can rotate in each frame interval. Since this method allows the simulation of lasers for each channel angle, we modified the Carla engine to accommodate channels of the following sensors: VLP-16, VLP-32c, HDL-64e. Each sensor has 16,32 and 64 channels. The Carla simulator calculates points' intensity using the system's attenuation and distance to point by the formula \ref{eqn:intensity-carla}. Here, $a$ refers to the attenuation of medium and $d$ refers to distance. 
\begin{equation}
\label{eqn:intensity-carla}
    I = e^{-a.d}
\end{equation}

\subsection{Simulation Results}
\begin{figure}[!ht]
    \centering
    \includegraphics[width=0.6\linewidth]{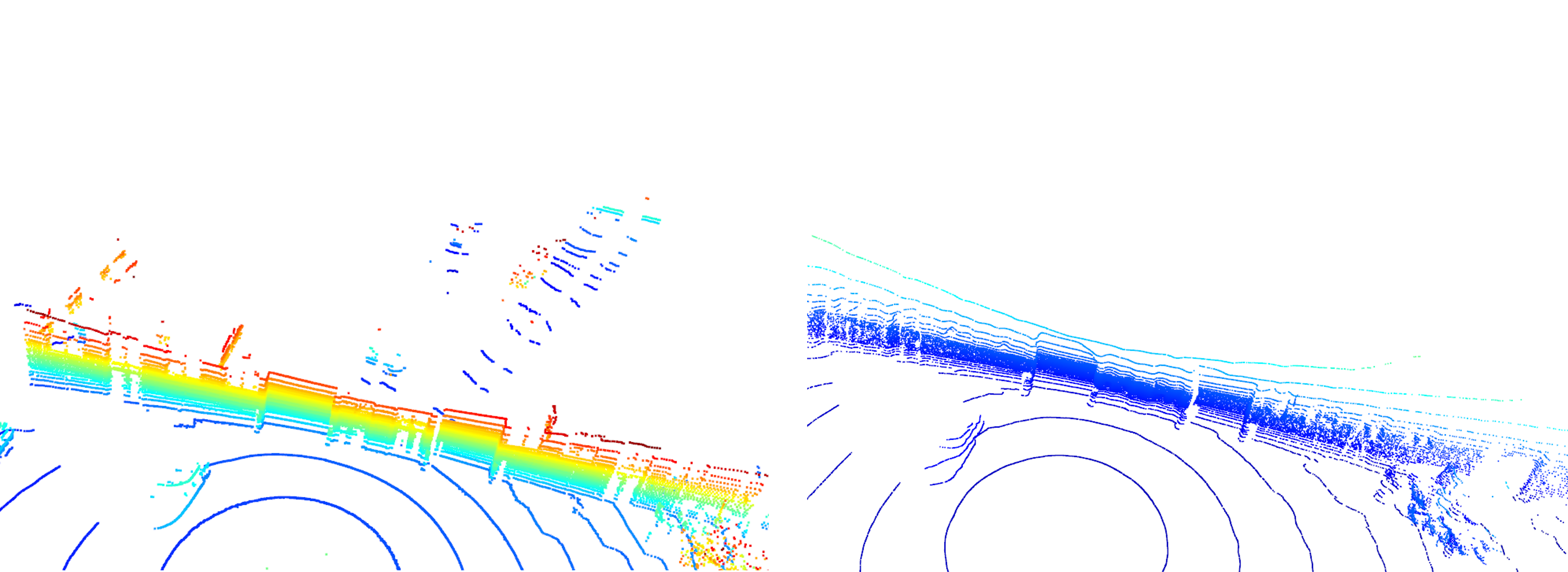}
    \caption{Visual Comparison of Scenarios with Windows. Real Scan (left) and Simulated  LiDAR Scan (Right)}
    \label{fig:realsim-glass}
\end{figure}

\begin{figure}[!ht]
    \centering
    \includegraphics[width=0.6\linewidth]{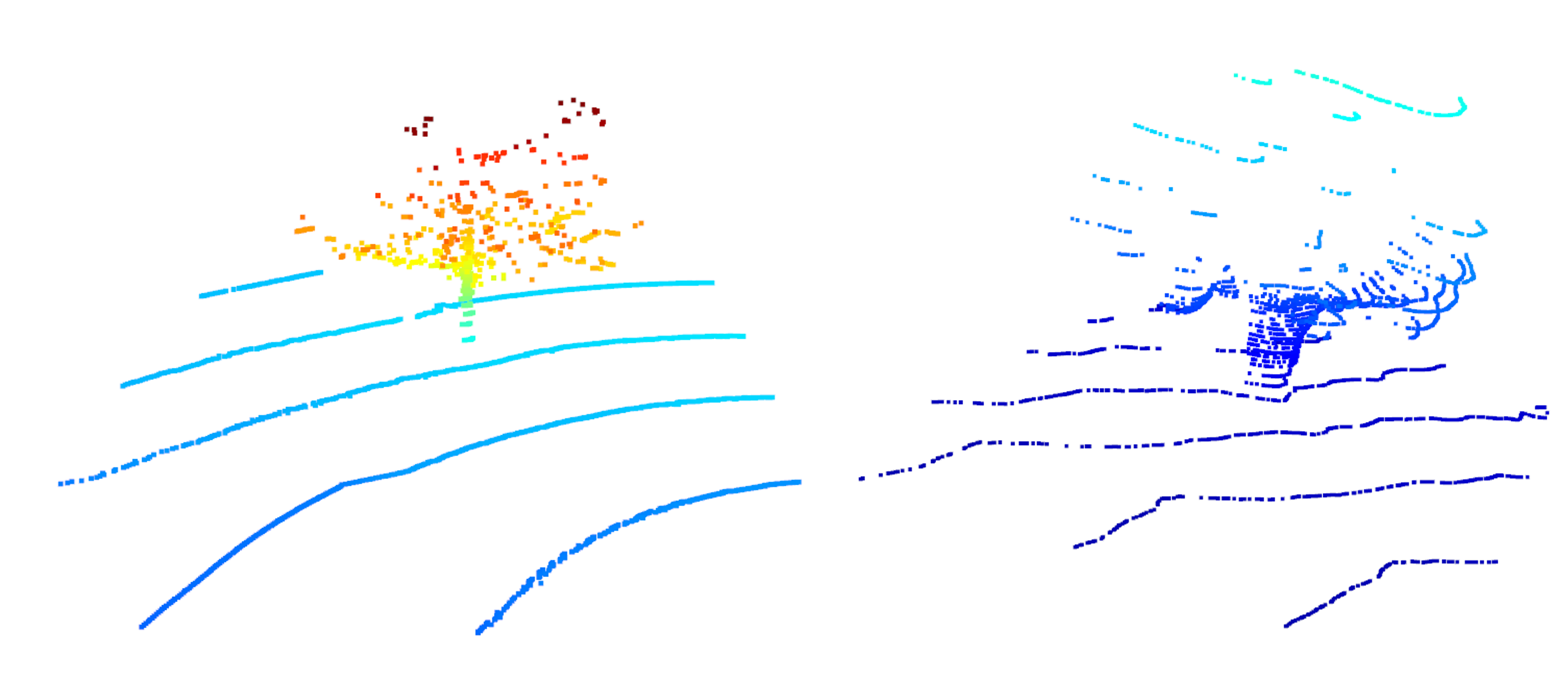}
    \caption{Visual Comparison of Scenarios with Trees. Real Scan (Left) and Simulated  LiDAR Scan (Right)}
    \label{fig:realsim-tree}
\end{figure}

\begin{figure}[!ht]
    \centering
    \includegraphics[width=0.6\linewidth]{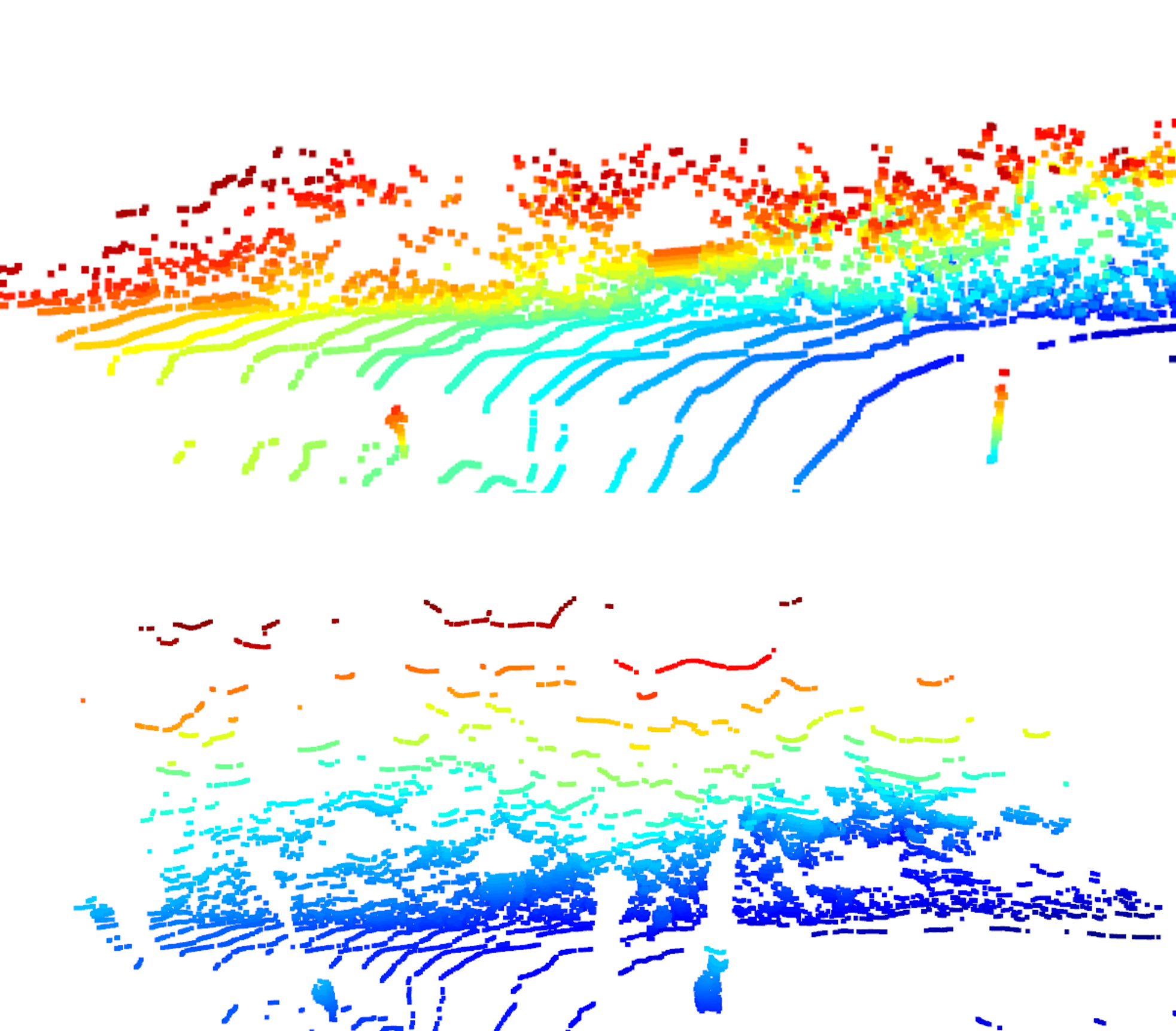}
    \caption{Visual Comparison of Scenarios with trees and bushes. Real Scan (Top) and Simulated  LiDAR Scan (Bottom)}
    \label{fig:realsim-bush}
\end{figure}

From the virtual testing environment, LiDAR scans were generated for similar odometers. Figure \ref{fig:realsim1-channels} visually compares Real and LiDAR scans. From visual inspection, the LiDAR scans have similar rings and geometry for ground elements. The simulated LiDAR scans and accurate LiDAR scans are different when simulating trees and materials such as glasses. One key reason is trees are difficult to map with point clouds, and generating a mesh for trees results in poor surface reconstruction. Also, since the approach for simulation only generates mesh and does not consider reflectivity and transparency of materials. Materials such as glasses are regarded as solid in simulation. In Figure \ref{fig:realsim-glass}, we see in the real scan the reflectance of glass produces a point cloud inside the wall, where we see no such outliers in the simulated scans. The trees in real LiDAR scans and Simulated LiDAR scans vary greatly; due to the mesh generation process, the trees act as a solid element in the VTE, whereas in the real world, the leaves scatter the LiDAR scans Figure \ref{fig:realsim-tree} shows a comparison of trees in both scenarios. In Figure \ref{fig:realsim-bush} we see LiDAR scan with bushes, the point cloud for the bushes in real world is much more sparse than simulated LiDARs. 
\ifhighlight
\else
\begin{figure*}[!htbp]
    \centering
    \includegraphics[width=1\linewidth]{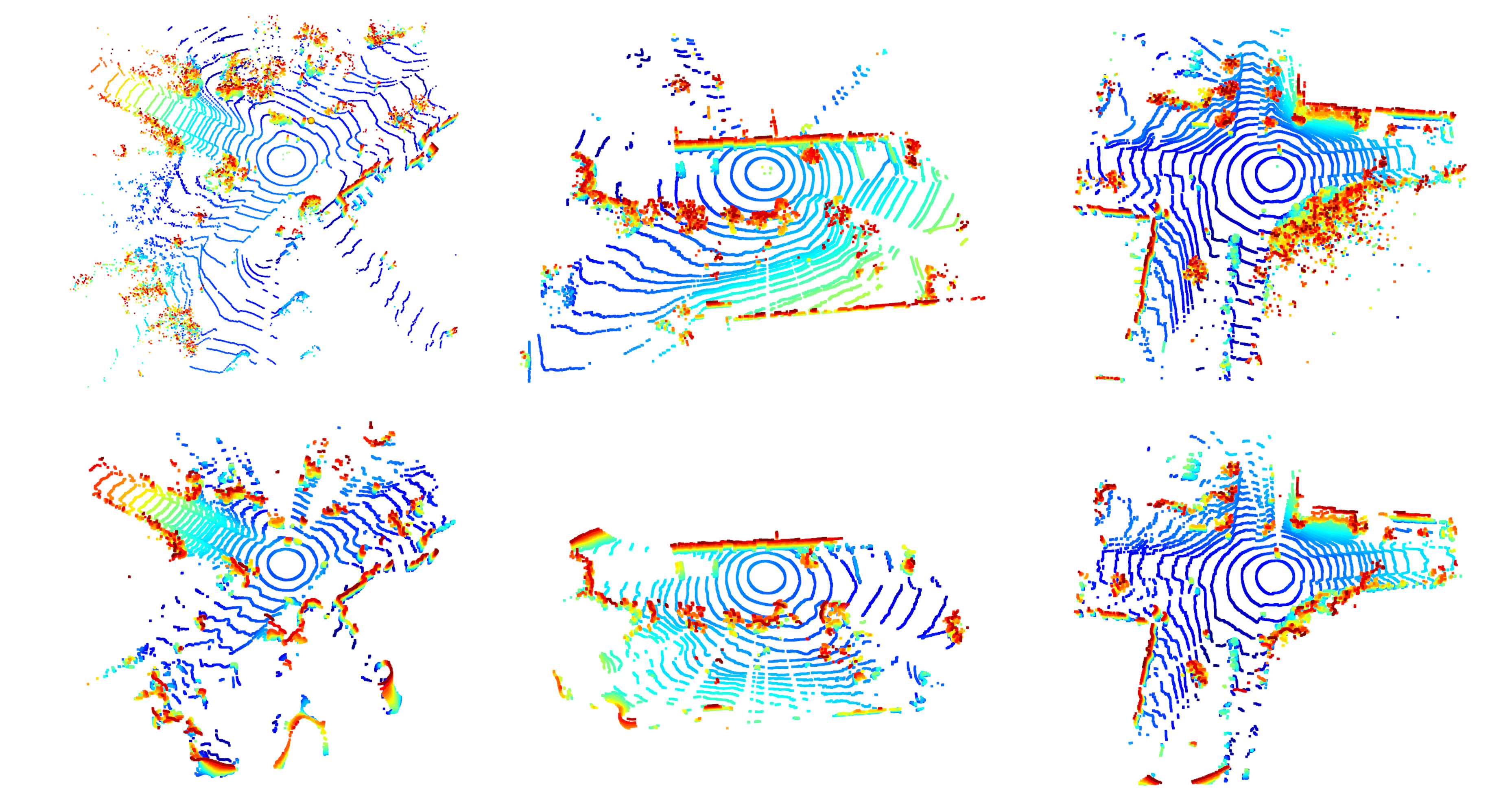}
    \caption{Visual Comparison of Real (Top) and Simulated (Bottom) LiDAR}
    \label{fig:realsim1-channels}
\end{figure*}
\fi

\ifhighlight
\else
\begin{figure}[!ht]
    \centering
    \includegraphics[width=0.6\linewidth]{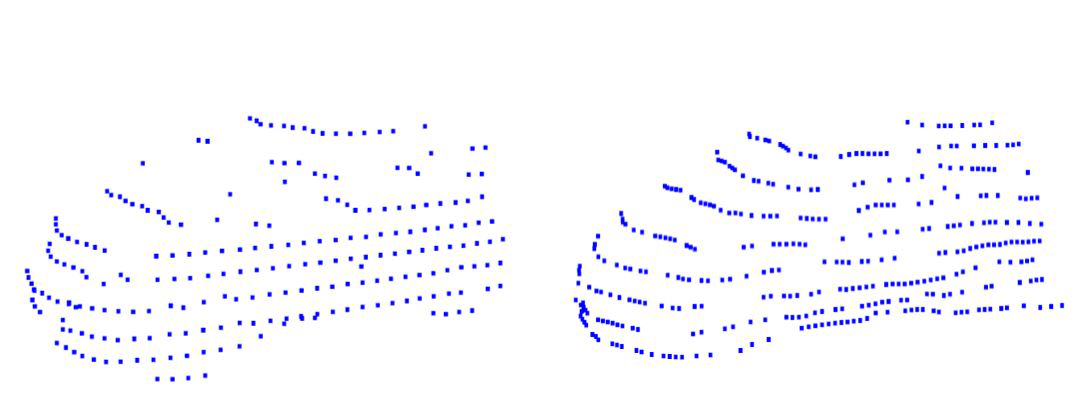}
    \caption{Visual Comparison of Scenarios with Cars. Real Scan (Left) and Simulated  LiDAR Scan (Right)}
    \label{fig:realsim-car}
\end{figure}
\fi

\subsubsection{Geometric Similarity}
\begin{figure}[!ht]
    \centering
    \includegraphics[width=1\linewidth]{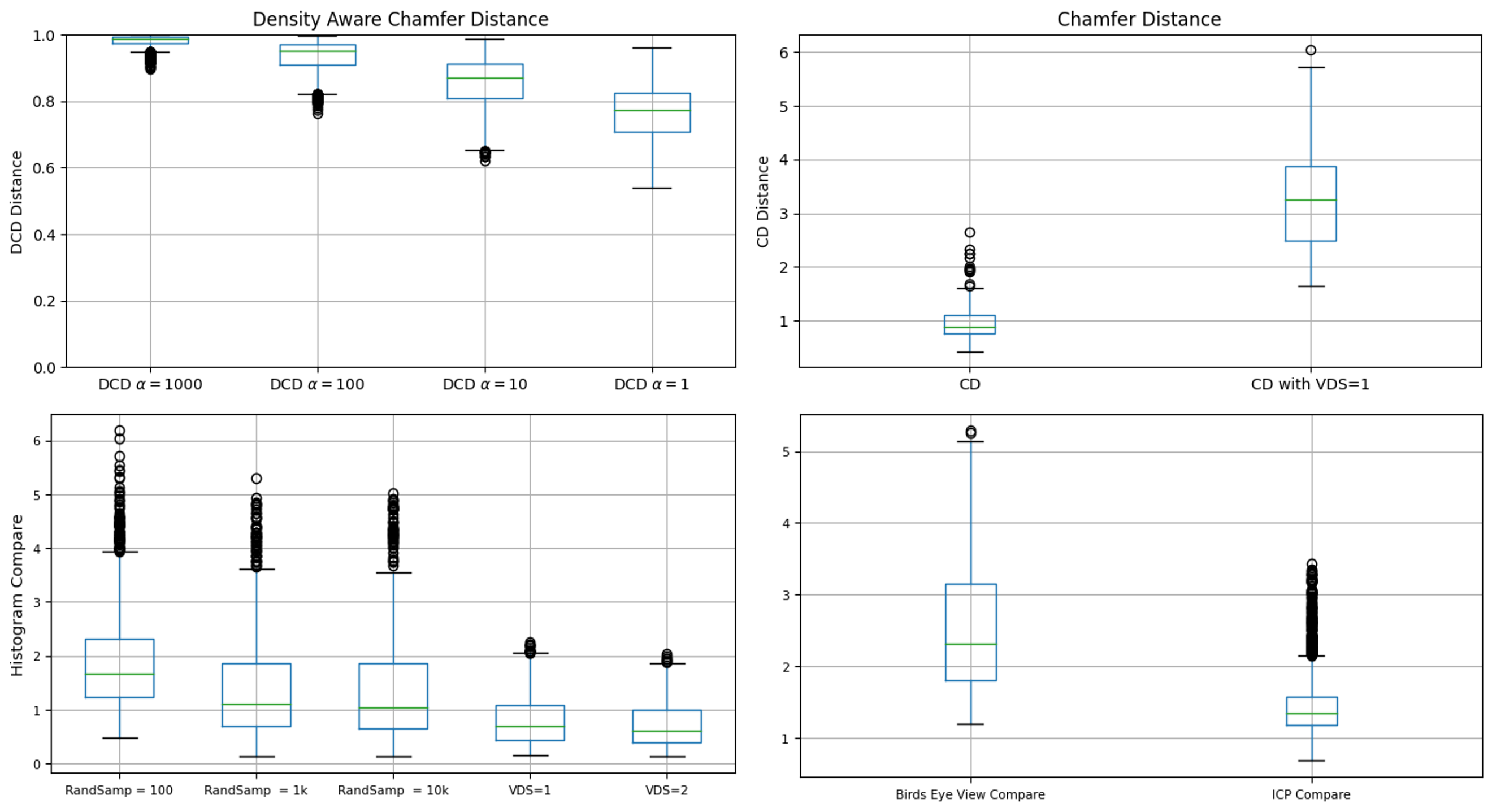}
    \caption{Distribution of different distance metrics when comparing simulated LiDAR and Real LiDAR of the same odometry. Top Left: Density Aware Chamfer Distances. Top Right: Chamfer Distance. Bottom Left; Histogram Methods. Bottom Left: Birds Eye View and ICP compare.}
    \label{fig:metric-geo-dist}
\end{figure}

The surveyed LiDAR scans were compared with their simulated twin with the same position and orientation in the VTE to set a benchmark score. With our approach of generating a virtual testing environment, the LiDAR scans have an average of $0.76$ Density Aware Chamfer Distance and $1.81$ Chamfer Distance. This suggests the LiDAR scans have a moderate level of geometric similarity. Table \ref{tab:metric-comparision-mean-std} contains comparison results for another similarity metric. In Figure \ref{fig:metric-geo-dist}, we show the distribution of different metrics for the comparison results. 

\begin{table}[!htbp]
\caption{Comparison results with different metrics}
\centering
\label{tab:metric-comparision-mean-std}
\begin{tabular}{|l|l|l|}
\hline
Metrics                                     & Mean     & STD      \\ \hline
Birds Eye View Compare & 2.55 & 0.90 \\ \hline
Chamfer Distance                            & 1.81 & 0.46 \\ \hline
Chamfer Distance Voxel Down Sample=1        & 3.23 & 0.83 \\ \hline
Histogram with  RnS 100 points  & 1.87 & 0.84 \\ \hline
Histogram with  RnS 1k points & 1.36 & 0.88 \\ \hline
Histogram with  RnS 10k points & 1.33 & 0.90 \\ \hline
Histogram with VDS 1 unit   & 0.80 & 0.44  \\ \hline
Histogram with VDS 2 unit   & 0.73  & 0.41 \\ \hline
Iterative Closest Point                     & 1.42 & 0.38 \\ \hline
Voxel Intersection Over Union               & 0.04 & 0.02 \\ \hline
Density Aware Chamfer Distance $\alpha=1k$   & 0.98 & 0.01 \\ \hline
Density Aware Chamfer Distance $\alpha=100$    & 0.93 & 0.04  \\ \hline
Density Aware Chamfer Distance $\alpha=10$      & 0.85 & 0.07 \\ \hline
Density Aware Chamfer Distance $\alpha=1$       & 0.76 & 0.08 \\ \hline
\end{tabular}
\end{table}

\subsubsection{Model Perception Comparison: Original LiDAR scans and Simulation output}
\begin{figure*}[!h]
    \centering
    \includegraphics[width=1\linewidth]{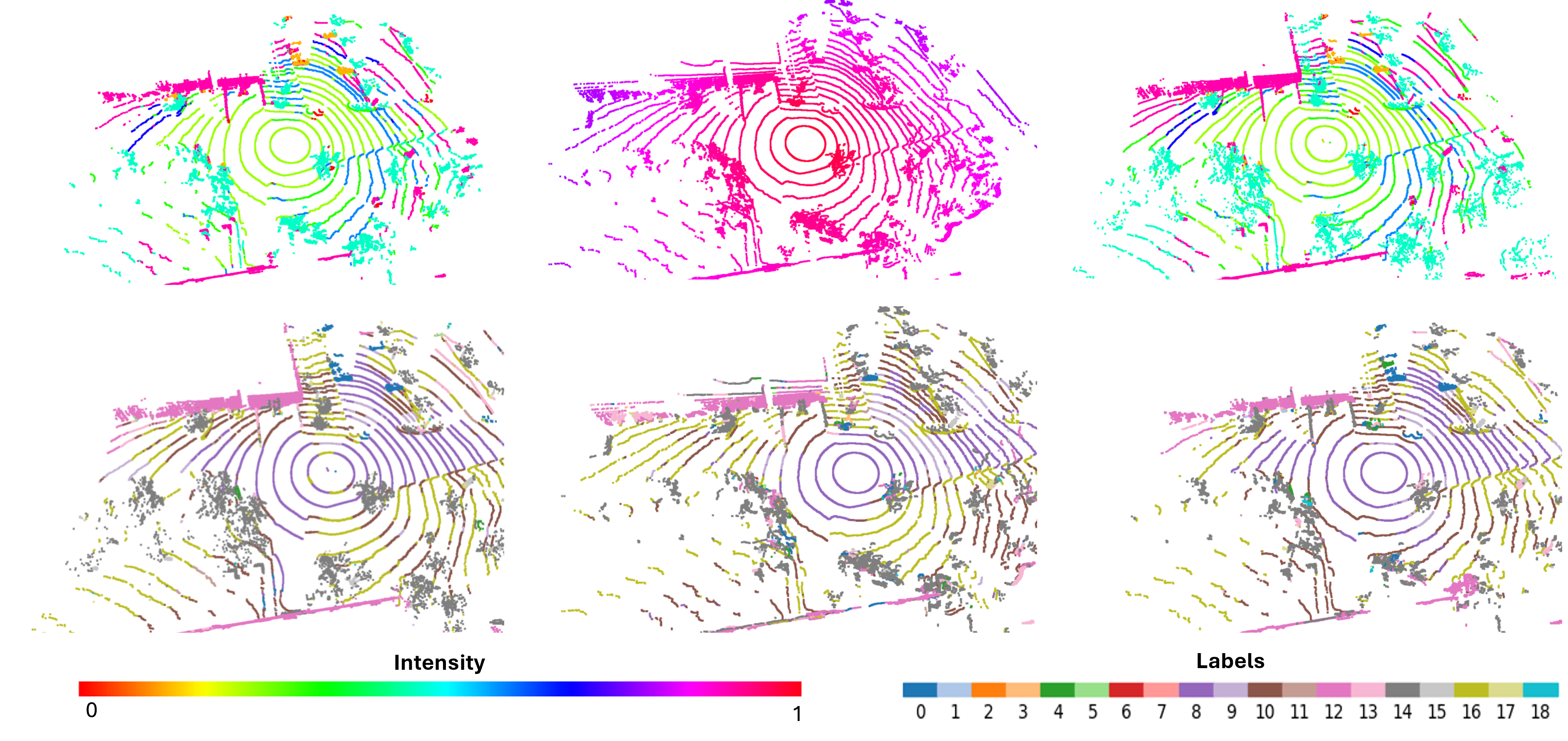}
    \caption{Intensity of LiDAR scans. Top Left: Intensity of Real LiDAR scan. Top Center: Intensity of simulated LiDAR scan. Top Right: Intensity of Simulated Scan with Corrected Intensity. Bottom Left: Segmentation output with Real LiDAR scan. Bottom Center: Segmentation output from simulated LiDAR scan. Segmentation Output from simulated LiDAR with corrected Intensity. The labels are as follows, 0: Car, 1: Bicycle, 2: Motorcycle, 3: Truck, 4: Other-vehicle, 5: Person, 6: Bicyclist, 7: Motorcyclist, 8: Road, 9: Parking, 10: Sidewalk, 11: Other-ground, 12: Building, 13: Fence, 14: Vegetation, 15: Trunk, 16: Terrain, 17: Pole, 18: Traffic-sign  }
    \label{fig:intensity}
\end{figure*}
To test model perception we compared the output of Sphereformer \cite{lai2023spherical} a LiDAR segmentation model pre-trained on Semantic Kitti Dataset. The model uses a Sparse Convolution Network-based U-Net architecture with transformer heads for semantic segmentation tasks. The model achieves a 74.8\% mIoU on the SemKITTI dataset. This model takes a single LiDAR scanned point cloud with intensity and input and generates semantic label data for each point. We use the model as a perception measure to check if real LiDAR scans and simulated LiDAR scans have similar perception results. Ideally, if the simulated LiDAR and original LiDAR are identical, the model should produce similar output. The comparison is by calculating the mean intersection by overlap (mIoU) of prediction output from the model. If both scans are identical, the mIoU should be 1. From the initial test with prediction LiDAR scans generated from the virtual testing environment and the original LiDAR scan of the same odometry, we see an average mIoU of $4.7\%$. This shows the detection model perceives both scans very differently. Although there is some level of geometric similarity, the intensity of both scans is very different. It can be assumed that if both scans have similar intensity, the model should have better similarity in terms of perception.

\subsubsection{Model Perception Comparison: Original LiDAR scans and Simulation output with corrected intensity}
To verify the assumption that if both the real and simulated LiDAR scans have similar intensity, the detection will improve. We corrected the simulation intensity by overlaying the intensity of real LiDAR scans on the simulated one. We used the nearest neighbour approach for each point in the simulated LiDAR scan and got the intensity from the closest point in the real LiDAR scan within a threshold of 1 unit.

\begin{figure}
    \centering
    \includegraphics[width=0.7\linewidth]{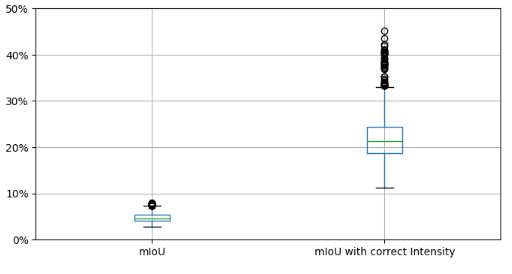}
    \caption{Mean Intersection over Union of prediction from Sphereformer model for real LiDAR scan and simulated LiDAR scan. The left boxplot shows the result for comparison with simulated intensity. The right box plot shows the result with corrected intensity for simulated scans. Here, the y-axis represents the mIoU values.}
    \label{fig:mious}
\end{figure}

By adopting this approach, we see an improvement in model perception from 4.7\% to 21.2\%. Figure \ref{fig:mious} shows the distribution of mIoU of detection between LiDAR scans from both sources. This result shows that the intensity of LiDAR scans influences model perception outputs. Figure \ref{fig:intensity} shows the intensity of the real, simulated LiDAR scan and intensity-corrected simulated LiDAR scan of the same odometry. The Figure shows the semantic segmentation output with the Sphereformer model.


\section{Conclusion}

\textcolor{black}{This paper proposes a method to compare simulated and real LiDAR scans using geometric similarity measures. We made pair-to-pair comparisons across real, augmented, and simulated LiDAR scans of different case studies. From the comparison of the case studies, we find that Density Aware Chamfer Distance is by far the most suitable geometric comparison method. \maybehl{Because it is sensitive to all modifiers, self-consistent, symmetrical, and most computationally efficient, and has a well-defined range of 0 to 1. Additionally, this metric contains a sensitivity variable $\alpha$ that allows for measuring the slightest differences in LiDAR scans, making it more robust across different scenarios} The sensitivity of the tested scenarios and computational efficiency determine the suitability. Second, we developed a LiDAR scan comparison methodology for comparing real and simulated LiDAR scans. The simulated copy of the LiDAR scans was generated using the CARLA simulator with exact sensor settings and pose. The simulated scans produced acceptable DCD values. Third, we compared the 3D object detection performance of the Sphereformer model using real, simulated LiDAR scans, and simulated LiDAR scans with corrected intensity. Our findings show that the 3D perception model detects similar intensities for corrected intensities.}

\textbf{Study Limitations.} The paper finds Density Aware Chamfer Distance a key similarity metric for comparing LiDAR scan pairs of the same odometry. Generating LiDAR scans in a real-world environment with the same odometry can pose a significant challenge. The metric only compares the geometry of point clouds, ignoring intensity. From the model perception comparison, intensity plays a vital role in model perception. Since our approach only converts 3D point clouds into a mesh, intensity is not considered. LiDAR intensity depends on factors such as medium attenuation, material reflectivity, and sensor settings. Generating a simulator that can accurately simulate LiDAR scans of diverse sensors is challenging. It is also difficult to validate the intensity of LiDAR scans. Another key limitation found in VTE development is that due to meshing, all the objects are not similar in real and simulated scans, such as vegetation reflecting materials. A standard geometric approach would not be the best way to validate these scans. Also, LiDAR scans in snow and rain generate scattered points in specific patterns. Evaluating these scatter patterns is also a challenge with Geometric approaches. 
Due to the lack of ground truth data, the model perception assessment compares the segmented output of the real and simulated scans. Assessing the accuracy of model perception for both data and ground truth data can open new insights into simulated model perception. Also, due to technical challenges, LiDAR scans with varying channels are only simulated in Carla. 

\textbf{Future Research.} With these limitations at hand, our future goal is to find evaluation metrics sensitive to geometry and other LiDAR properties, such as intensities. Digital twins can be developed with an advanced approach, such as using neural kernel fields to generate the mesh. We can also generate ground truth labelled data to better evaluate model perceptions. In future work, we will use 128-channel LiDAR to map real-world environments to generate a more accurate digital twin with advanced mesh generation methods.

\bibliographystyle{unsrt}  
\bibliography{references}  

\end{document}